\documentclass[5p]{elsarticle}

\journal{Pattern Recognition}

\usepackage[utf8]{inputenc}         
\usepackage[T1]{fontenc}            
\usepackage{url}                    
\usepackage{booktabs}               
\usepackage{amsfonts}               
\usepackage{nicefrac}               
\usepackage{microtype}              

\usepackage{float}                  
\usepackage{subcaption}             
\usepackage{adjustbox}
\usepackage{xspace}
\usepackage{setspace}

\usepackage{amsmath}
\usepackage{amssymb}
\usepackage{mathtools}              
\usepackage[noline,ruled]{algorithm2e}
\SetAlgoInsideSkip{medskip}
\usepackage{enumitem}               
\usepackage{array}
\usepackage{longtable}
\usepackage{multirow}
\usepackage{threeparttable}
\usepackage{makecell}

\usepackage{times}                  
\usepackage{soulutf8}               
\usepackage{hyperref}               

\setuldepth{6.3~~($-$6.3)}          

\DeclareRobustCommand{\ie}{i.e.\@\xspace}

\makeatletter
\DeclareRobustCommand{\etc}{%
    \@ifnextchar{.}%
    {etc}%
    {etc.\@\xspace}%
}

\DeclareRobustCommand\onedot{\futurelet\@let@token\@onedot}
\def\@onedot{\ifx\@let@token.\else.\null\fi\xspace}
\def\etal{\emph{et al}\onedot}

\makeatother

\newcommand{\gph}[2]{\includegraphics[width=#1\linewidth]{#2}}

\newcommand{\bft}[1]{\textbf{#1}}
\newcommand{\itt}[1]{\textit{#1}}

\newcolumntype{L}[1]{>{\raggedright\arraybackslash}p{#1}}   
\newcolumntype{C}[1]{>{\centering\arraybackslash}p{#1}}     
\newcolumntype{`}{>{\global\let\currentrowstyle\relax}}     
\newcolumntype{~}{>{\currentrowstyle}}                      

\newcommand{\rowstyle}[1]{\gdef\currentrowstyle{#1}#1\ignorespaces }
\newcommand{\rbf}{\rowstyle{\bfseries} \boldmath}           

\newcommand{\fmtr}[2]{\multirow{#1}{*}{#2}}

\DeclareMathOperator{\sample}{z}
\DeclareMathOperator{\bern}{Bern}
\DeclareMathOperator{\round}{Round}
\DeclareMathOperator{\sigmoid}{\sigma}

\DeclarePairedDelimiter\abs{\lvert}{\rvert}%
\DeclarePairedDelimiter\norm{\lVert}{\rVert}%
\DeclarePairedDelimiter\set\{\}%

\newcommand{\NNZ}{p_{nnz}}
\newcommand{\TotalParams}{p_{total}}

\begin{document}

\begin{frontmatter}

\title{End-to-End Supermask Pruning:\\Learning to Prune Image Captioning Models}

\author[fsktm]{Jia~Huei~Tan}
\ead{tanjiahuei@gmail.com}
\address[fsktm]{Center of Image and Signal Processing (CISiP), Department of Artificial Intelligence, \\ Universiti Malaya, 50603 Kuala Lumpur, Malaysia}

\author[fsktm]{Chee~Seng~Chan\texorpdfstring{\corref{corrauthor}}{*}}
\cortext[corrauthor]{Corresponding author}
\ead{cs.chan@um.edu.my}

\author[engine]{Joon~Huang~Chuah}
\ead{jhchuah@um.edu.my}
\address[engine]{Department of Electrical Engineering, Universiti Malaya, 50603 Kuala Lumpur, Malaysia}

\begin{abstract}

    With the advancement of deep models, research work on image captioning has led to a remarkable gain in raw performance over the last decade, along with increasing model complexity and computational cost. However, surprisingly works on compression of deep networks for image captioning task has received little to no attention.
    For the first time in image captioning research, we provide an extensive comparison of various unstructured weight pruning methods on three different popular image captioning architectures, namely \itt{Soft-Attention}, \itt{Up-Down} and \itt{Object Relation Transformer}.
    Following this, we propose a novel end-to-end weight pruning method that performs gradual sparsification based on weight sensitivity to the training loss.
    The pruning schemes are then extended with encoder pruning, where we show that conducting both decoder pruning and training simultaneously prior to the encoder pruning provides good overall performance.
    Empirically, we show that an 80\% to 95\% sparse network (up to 75\% reduction in model size) can either match or outperform its dense counterpart. The code and pre-trained models for \itt{Up-Down} and \itt{Object Relation Transformer} that are capable of achieving CIDEr scores $>$120 on the MS-COCO dataset but with only 8.7 MB and 14.5 MB in model size (size reduction of 96\% and 94\% respectively against dense versions) are publicly available at \url{https://github.com/jiahuei/sparse-image-captioning}.

\end{abstract}


\end{frontmatter}



\section{Introduction}
\label{sec: Pruning: Introduction}

Over the past decade, continuous research on image captioning using deep neural networks (DNNs) has led to a steady improvement in the overall model performance. For instance, CIDEr\footnote{Consensus-based Image Description Evaluation (CIDEr) is a widely-used metric for caption quality by measuring the level of consensus between generated captions and ground-truth captions.} scores \cite{vedantam2015cider} of state-of-the-art (SOTA) models have doubled from 66 points \cite{karpathy2015deep} to 130 points and beyond \cite{herdade2019image,cornia2020meshed} recently on the MS-COCO dataset \cite{lin2014microsoft}. However, such gains are usually achieved at the expense of model size using heavily parameterised models, where the decoder size had quadrupled from 12 million \cite{xu2015show} to 55 million \cite{herdade2019image} parameters (see Table~\ref{table: Pruning: MS-COCO SOTA} for details).

In an effort to reduce model size, various pruning techniques have been proposed to remove unimportant weights from a network. Generally, there are multitudes of benefits to be gained from weight pruning: it provides opportunities for improvements in terms of i) \itt{speed}, ii) \itt{storage}, and iii) \itt{energy consumption}, especially during the deployment stage. For \itt{speed}, highly-sparse models with significantly fewer non-zero parameters can enjoy faster run-times when combined with efficient SpMM kernels \cite{elsen2020fast,wang2020sparsert}. This is particularly true for Recurrent Neural Network (RNN) and Transformer whose matrix-multiplication computations are bottlenecked by bandwidth \cite{kalchbrenner2018efficient}. For \itt{storage}, compressed models are easier to be deployed onto mobile devices. Moreover, compressing SOTA model checkpoints into tens of MB can potentially accelerate the dissemination of research findings, result reproduction and experimentation. Finally, for \itt{energy consumption}, small RNN kernels produced via pruning can be stored in on-chip SRAM cache with lower energy requirements rather than DRAM memory \cite{han2015learning}, reducing carbon footprint.

While there is no shortage of pruning methods for image classification and translation tasks \cite{zhu2017prune,chirkova2018bayesian,louizos2018learning,lee2018snip} (see Sec.~\ref{sec: Pruning: Related Works} for more), their applicability to multimodal contexts such as image captioning is still under-explored. To the best of our knowledge, there is only one prior work that involved pruning an image captioning model, and that is by Dai \etal{} \cite{dai2020grow}. We hypothesise that this lack of progress is due to several difficulties. Firstly, weights are shared and reused across time steps, complicating variational pruning methods proposed for feed-forward networks \cite{chirkova2018bayesian}. Secondly, naively performing structured pruning on Long-Short Term Memory (LSTM) kernels can lead to invalid units due to inconsistent dimensions \cite{wen2018learning}. Thirdly, whereas small CIFAR datasets allow quick experimentation and iteration for Convolutional Neural Network (CNN) pruning \cite{louizos2018learning,frankle2018lottery}, there is a lack of an equivalent dataset in image captioning. Finally, image captioning is an inherently complex multimodal task; thus any proposed method must be able to perform well on both image and language domains.

\begin{figure}[t]
    \centering    
    \gph{1.0}{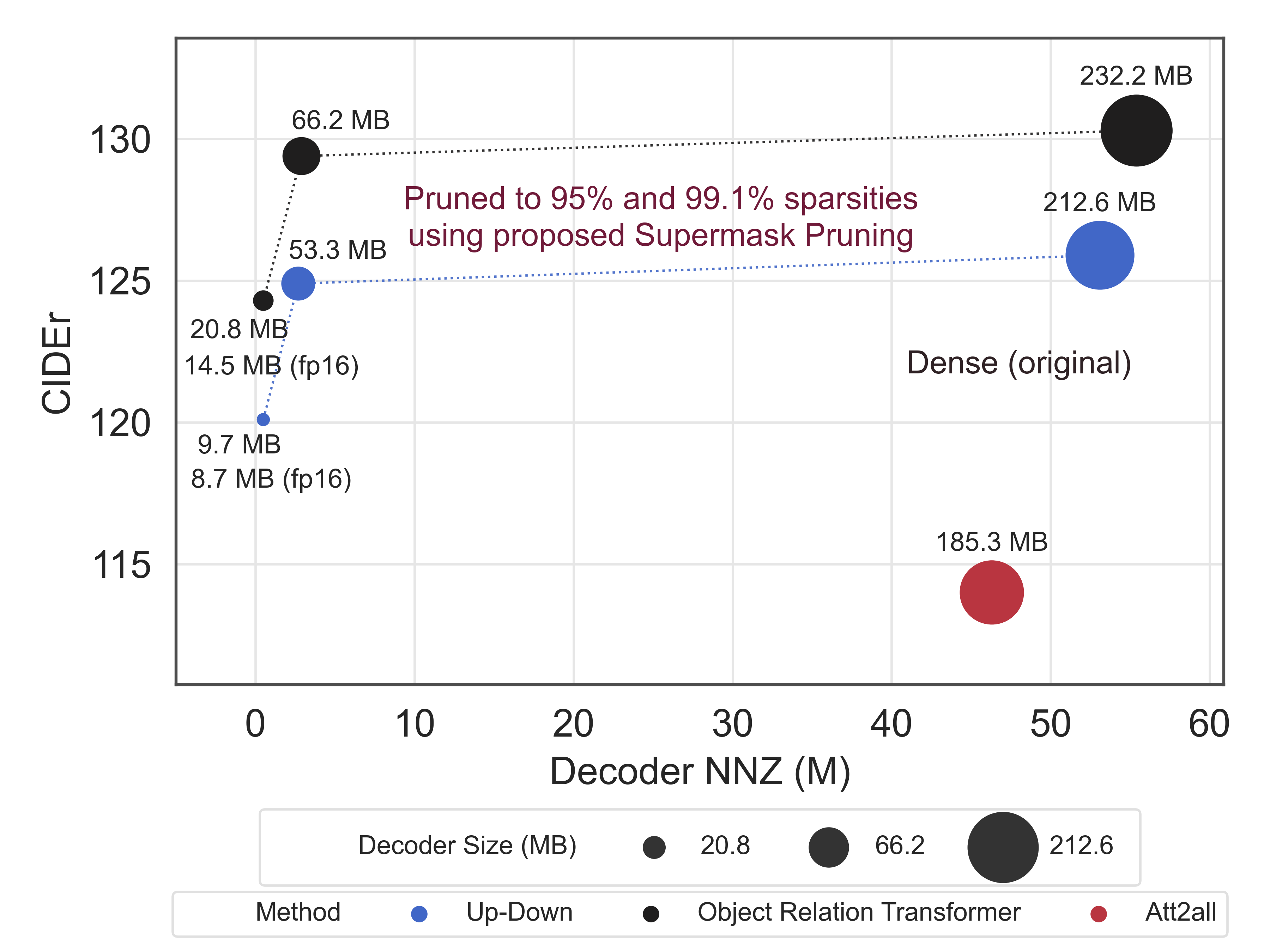}
    \caption{Our proposed Supermask Pruning (SMP) method can produce 99.1\% sparse networks that are capable of achieving CIDEr scores of 120 and above on the MS-COCO dataset, see Table~\ref{table: Pruning: MS-COCO SOTA}.}
    \label{fig: Pruning: Teaser}
\end{figure}

To this end, this paper attempts to answer the following questions:
\begin{enumerate}[label=\arabic*)]
    \item {Which weight pruning method produces the best results on image captioning models?}
    \item {Is there an ideal sparsity where a sparse model can match or even outperform its dense counterpart?}
    \item {What is the ideal prune-finetune sequence for pruning both the pre-trained encoder and decoder?}
    \item {Can a sparse captioning model outperform a smaller but dense model?}
\end{enumerate}
\noindent with an extensive comparison of various unstructured weight pruning methods on three different SOTA image captioning architectures, namely \itt{Soft-Attention} (SA) \cite{xu2015show}, \itt{Up-Down} (UD) \cite{anderson2018bottom} and \itt{Object Relation Transformer} (ORT) \cite{herdade2019image}. Following this, we propose a novel end-to-end weight pruning method that performs gradual sparsification while maintaining the overall model performance. The pruning schemes are then extended with encoder pruning, where several prune-finetune sequences are explored. Empirically, we show that conducting both decoder pruning and training simultaneously prior to the encoder pruning-and-finetuning provides better raw performance. Also, we show that for a given performance level, a large-sparse LSTM captioning model is better than a small-dense one in terms of model costs. 

As a summary, the core contributions of this paper are threefold. Firstly, this is the first extensive attempt at exploring unstructured model pruning for image captioning task. Empirically, we show that 80\% to 95\% sparse networks can either match or even slightly outperform their dense counterparts (Sec.~\ref{subsec: Pruning: Experiments: Pruning}). In addition, we propose a pruning method -- Supermask Pruning (SMP) that performs continuous and gradual sparsification during training stage based on parameter sensitivity in an end-to-end fashion. Secondly, we investigate an ideal way to combine pruning with fine-tuning of pre-trained CNN, and show that both decoder pruning and training should be done before pruning the encoder (Sec.~\ref{subsec: Pruning: Experiments: Pruning Sequence}). Finally, we release the pre-trained sparse models for UD and ORT that are capable of achieving CIDEr scores $>$120 on the MS-COCO dataset; yet are only 8.7 MB (reduction of 96\% compared to dense UD) and 14.5 MB (reduction of 94\% compared to dense ORT) in model size (Fig.~\ref{fig: Pruning: Teaser} and Sec.~\ref{subsec: Pruning: Experiments: SOTA}). Our code and pre-trained models are publicly available\footnote{\url{https://github.com/jiahuei/sparse-image-captioning}}.


\section{Related Works}
\label{sec: Pruning: Related Works}

\subsection{Image Captioning}
\label{subsec: Pruning: Related: Image Captioning}

Since the advent of deep neural networks, research on image captioning can be characterised by numerous architectural innovations in pursuit of raw performance (see \cite{hossain2019comprehensive} for a complete survey). 
The first major innovation came in the form of an end-to-end captioning network that directly generates a caption given an image \cite{karpathy2015deep,tan2019phrase}. 
Next came visual attention, in which one or more CNN feature maps were used to guide and condition the caption generation process \cite{xu2015show,fu2018image}. 
There are also numerous works that used attributes as a way to directly inject salient information into the decoder \cite{chen2018show,ding2019neural,ji2021divergent}. 
Following that, \cite{anderson2018bottom} employed an object detector to generate image features as a form of hard-attention; which along with Transformer, became a popular captioning paradigm \cite{herdade2019image,cornia2020meshed,wang2020word}. 
Concurrently, substantial effort has been put into reinforcement learning which allowed non-differentiable caption metrics to be used as optimisation objectives \cite{rennie2017self,chen2018temporal}.
While these methods have been successful in advancing the SOTA performance, minimal effort has been made on reducing model cost \cite{para2017exp,tan2019comic}, which is the main motivation of this work.

\subsection{Structured or Channel Pruning}
\label{subsec: Pruning: Related: Structured Pruning}

Structured pruning is a coarse-grain pruning technique whereby entire rows, columns or channels of fully connected or convolutional weights are removed.
There are extensive prior work in this direction targeted at feed-forward CNNs, including \cite{luo2020autopruner,zhuang2018discrimination,lin2020hrank,li2020eagleeye,lin2020channel} just to name a few. 
At the same time, structured pruning of RNNs is also widely explored \cite{wen2018learning,yu2019learning,wen2020structured}.

Since structured pruning reduces model dimensions, the resulting network is more amenable to run-time speed-up. However, this advantage comes with several costs:
\textbf{(a) Architectural constraints:} For gated RNNs such as LSTM, structured pruning requires that the pruned rows and columns of the recurrent weight kernels be aligned with each other; otherwise it may lead to invalid units \cite{wen2018learning}. The same is true for attention kernels, which is extensively used in modern captioning architectures.
\textbf{(b) Lower sparsity:} Structured pruning usually provides lower sparsity for a given performance loss \cite{crowley2018pruning,liu2019rethinking}, often in the range of 40\% (1.7$\times$) to 90\% (10$\times$). In contrast, we demonstrate that unstructured pruning can prune an order of magnitude more at 99\% (100$\times$) while maintaining performance (see Fig.~\ref{fig: Pruning: UpDown ORT COCO}).

\subsection{Unstructured Pruning}
\label{subsec: Pruning: Related: Unstructured Pruning}

Recently, unstructured pruning has enjoyed emerging support, including Fast SpMM kernels \cite{kalchbrenner2018efficient,elsen2020fast,wang2020sparsert} and block-sparsity support by NVIDIA Ampere GPU\footnote{\url{https://developer.nvidia.com/blog/nvidia-ampere-architecture-in-depth}} and HuggingFace Transformers library\footnote{\url{https://github.com/huggingface/pytorch_block_sparse}}. While there exist numerous unstructured pruning methods \cite{chirkova2018bayesian,louizos2018learning,wang2019eigendamage}, we focus on methods applied to RNN and NLP models with the following characteristics: \bft{(a) Straightforward:} Reasonably simple to implement and integrate into a standard deep network training workflow. \bft{(b) Effective:} Able to prune at least 80\% of parameters without compromising performance. \bft{(c) Efficient:} Does not require expensive iterative pruning and retraining cycles. 
Thus, we arrive at the following pruning methods as a solid starting point for exploring image captioning model pruning:

\sloppy 

\textbf{(1) Hard / one-shot magnitude-based pruning:} \cite{see2016compression} first investigated three magnitude-based schemes for translation model with multi-layer LSTM, namely class-blind, class-uniform and class-distribution. \textit{Class-blind} removes parameters with the smallest absolute value regardless of weight class. In contrast, \textit{class-uniform} prunes every layer to the same sparsity level. \textit{Class-distribution} \cite{han2015learning} prunes parameters smaller than a global factor of the class standard deviation. Experiments found that class-blind produced the best results. 

\fussy

\textbf{(2) Gradual magnitude-based pruning:} First introduced by \cite{narang2017exploring} to prune parameters gradually over the course of training, it was extended by \cite{zhu2017prune} via a simplified pruning curve with reduced hyperparameters. The simplified slope has a single phase, governed by a cubic function that determines the sparsity level at each training step. Their method is tested on deep CNN, stacked LSTM and GNMT models on classification, language modelling and translation. 

\textbf{(3) SNIP:} \cite{lee2018snip} proposed a saliency criterion for identifying structurally important connections. The criterion is computed as the absolute magnitude of the derivative of training loss with respect to a set of multiplicative pruning masks. Guided by the saliency criterion, single-shot pruning of CNN and RNN were performed at initialisation, prior to training. It is evaluated on the task of image classification.

\textbf{(4) Lottery ticket (LT):} It is a seminal work by \cite{frankle2018lottery} which put forth ``The Lottery Ticket Hypothesis''. It states that there exists a subnetwork in a randomly initialised dense neural network, such that when trained in isolation can match the test accuracy of the original network. By iteratively pruning and resetting networks to their original initialisation values, the authors found sparse networks that can reach the original dense accuracy within equal or shorter training iterations. It is tested on CNNs for image classification.

\textbf{(5) Supermask:} The work by \cite{zhou2019deconstructing} explored various aspects of Lottery Tickets in order to determine the reason behind its success. In the process, the authors discovered that binary pruning masks can be learned in an end-to-end fashion. However as formulated, only the masks were optimised, and there is no straightforward way to control network sparsity. Another work by \cite{srinivas2017training} optimised both masks and weights of CNNs, yet similarly, its final sparsity is influenced indirectly via a set of regularisation hyperparameters.
In this work, we extend \itt{Supermask} with a \textbf{novel sparsity loss} (Eq.~\eqref{eq: Pruning: sparsity loss}) to directly control the final sparsity of the network.

Among the prior works on model pruning, only the work by Dai \etal{} \cite{dai2020grow} involved image captioning. However, there exists several differences with our work: 
\textbf{(a)} only the H-LSTM cell is pruned;
\textbf{(b)} CNN weight pruning is not investigated;
\textbf{(c)} grow-and-prune (GP) method \cite{dai2019nest} used requires expensive and time-consuming ``grow'' and ``prune-retrain'' cycles. 
In contrast, our approach prunes both encoder and decoder in-parallel with regular training. Nevertheless, we provide a performance comparison with H-LSTM in Table~\ref{table: Pruning: H-LSTM}.


\section{Supermask Revisited}
\label{sec: Pruning: Supermask}

Supermask \cite{zhou2019deconstructing} is a network training method proposed by Zhou \etal{} as part of their work on studying the Lottery Tickets phenomenon \cite{frankle2018lottery}. Their work aimed to uncover the critical elements that contributed towards the good performance of ``winning tickets'': sparse networks that emerged from iterative prune-reset cycles. In the process, the authors discovered that an untrained, randomly initialised network could attain test performance that is significantly better than chance. This is achieved by applying a set of well-chosen masks to the network weights, effectively pruning it. These masks were hence named ``Supermask'', in that they are able to boost performance even without training of the underlying weights.

\subsection{Learning Supermasks}
\label{subsec: Pruning: Supermask: Learning}

Supermasks are learned in an end-to-end fashion via stochastic gradient descent (SGD). For every weight matrix $W$ to be pruned, a gating matrix $G$ with the same shape as $W$ is created. This gating matrix $G$ operates as a masking mechanism that determines which of the parameter $w \in W$ will be involved in both the forward-execution and back-propagation of the graph. For a model with $R$ layers, we now have two sets of parameters: gating parameters $\phi = \set{G_{1:R}}$ and network parameters $\theta = \set{W_{1:R}, B_{1:R}}$. To this end, the effective weight tensor $W^{\prime}$ is computed following Eq.~\eqref{eq: Pruning: weight masking}:

\begin{equation}\label{eq: Pruning: weight masking}
    W^{\prime} = W \odot G^{b}
\end{equation}

\noindent where $W , G \in\mathbb{R}^{D}$ are the original weight and gating matrices with shape $D$; and superscript $\left(\cdot\right)^{b}$ indicates binary variables. $\odot$ is element-wise multiplication.

In order to achieve the desired masking effect, $G^{b}$ must contain only ``hard'' binary values, \ie{} $G^{b} \in \set{0, 1}^D$. Therefore, matrix $G$ containing continuous values is transformed into binary matrix $G^{b}$ using a composite function $\sample\left( \sigmoid\left( \cdot \right) \right)$. Here, $\sigmoid(\cdot)$ is a point-wise function that squeezes continuous values into the interval $(0, 1)$; whereas $\sample(\cdot)$ is another point-wise function that samples from the output of $\sigmoid(\cdot)$. This is shown in Eq.~\eqref{eq: Pruning: gating sample}:

\begin{equation}\label{eq: Pruning: gating sample}
    G^{b} = \sample\left( \sigmoid\left( G \right) \right)
\end{equation}

Sampling from $\sigmoid\left( G \right)$ is done by treating $\sigmoid\left( G \right)$ as Bernoulli random variables, and then performing an ``unbiased draw''. Unbiased draw is the sampling process where each gating value $g \in [0, 1]$ is binarised to $1$ with probability $g$ and $0$ otherwise, \ie{} $\sample(\cdot) = \bern(\cdot)$. Sigmoid function is employed as $\sigmoid(\cdot)$. Finally, the effective weight $W^{\prime}$ can be computed as follows by modifying Eq.~\eqref{eq: Pruning: weight masking}:

\begin{equation}\label{eq: Pruning: weight masking bern}
    W^{\prime} = W \odot \bern( \sigmoid( G ) )
\end{equation}

Before training, all the gating variables $\phi = \set{G_{1:R}}$ are initialised with the same constant value $m$, whereas the weights of the network are initialised randomly. The authors found that the utilisation of $\bern(\cdot)$ helped to mitigate the bias arising from the constant value initialisation by injecting stochasticity into the training process.

Although Supermask is an effective pruning technique, the formulation as presented does not allow for easy control of final network sparsity. Instead, the pruning ratios were indirectly controlled via the pruning mask initialisation magnitude. In order to address this limitation, we proposed Supermask Pruning (SMP) method that is explained next.


\section{Supermask Pruning (SMP)}
\label{sec: Pruning: Proposed}

\begin{figure}[t]
    \centering    
    \gph{0.8}{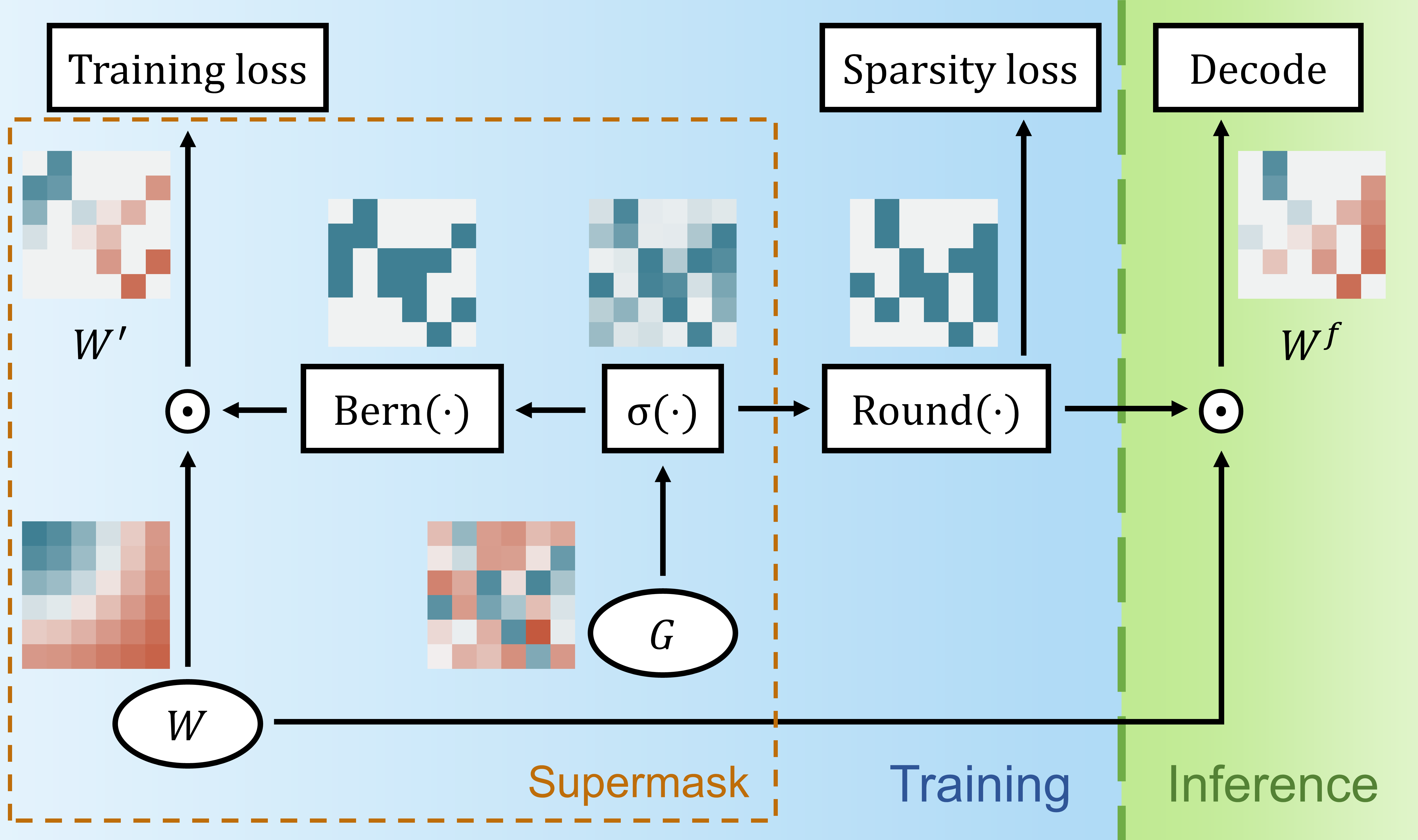}
    \caption{An overview of our proposed Supermask Pruning (SMP). During the training stage, each weight is probabilistically retained via Bernoulli sampling on the gating parameters which are sparsified via a sparsity loss. Upon training completion, all weights are sparsified by multiplying with binarised gating parameters.}
    \label{fig: Pruning: Overview}
\end{figure}

In this paper, we propose a simple yet effective method to directly control the final weight sparsity of models pruned based on the Supermask framework. To achieve this, a \emph{novel sparsity loss} $L_{s}$ is formulated which allows one to drive the sparsity level of gating variables $\phi$ to a user-specified level $s_{target}$. We name our method Supermask Pruning (SMP), and an overview is illustrated in Fig.~\ref{fig: Pruning: Overview}. The complete algorithm is given in Algorithm~\ref{algo: Pruning: Supermask Pruning}.

\subsection{Sparsity Loss}
\label{subsec: Pruning: Proposed: Sparsity Loss}

Technically, a straightforward way to influence the sparsity and pruning rate of Supermask is to introduce an $L_1$ regularisation term as follows:
\begin{equation}\label{eq: Pruning: sparsity loss naive}
    L_{s} = \abs*{ \: s_{target} - \left( 1 - \frac{\NNZ}{\TotalParams} \right) \: }
\end{equation}
\noindent where $\NNZ$ is the number of non-zero (NNZ) gating parameters; $\TotalParams$ is the total number of gating parameters; $n$ and $n_{max}$ are the current and final training step respectively. Such a regularisation term as formulated in Eq.~\eqref{eq: Pruning: sparsity loss naive} would apply a downward pressure on the magnitude of the gating parameters $G$ over the course of training, so that by the end of training, most of the gating parameters would have magnitudes smaller than zero. Ideally, these negative-valued gating parameters would represent weights that are least important, and can thus be removed without significant performance impact. At the same time, smaller gating magnitudes will cause more weights to be dropped more frequently, which in turn would allow the network to learn to depend on fewer parameters.

However, while naively applying the regularisation term (Eq.~\eqref{eq: Pruning: sparsity loss naive}) can produce networks with the desired sparsities, it does not achieve optimal performance. Our preliminary experiments found that constant application of $L_{s}$ causes weights to be dropped too early in the training process. In other words, it leads to an over-aggressive pruning schedule. To mitigate this, we propose to perform loss annealing by adding a variable weight $\alpha$ to the $L_{s}$ term in the cost function. 

Our idea is at the beginning of training, the value of $\alpha$ is set to zero to allow network learning to progress without any pruning being done. As training progresses, the value of $\alpha$ is gradually increased, forcing the model towards a sparse solution. Specifically, this loss annealing is done using an inverted cosine curve. Our experiments found that such gradual weight pruning produces better results, which is consistent with the observations in \cite{narang2017exploring,zhu2017prune}. Thus, our final sparsity loss $L_{s}$ is given as:
\begin{equation}\label{eq: Pruning: sparsity anneal}
    \alpha = 1 - \frac{1}{2} \left( 1 + \cos \left( \frac{n \pi}{n_{max}} \right) \right)
\end{equation}
\begin{equation}\label{eq: Pruning: sparsity loss}
    L_{s} = \alpha \: \abs*{ \: s_{target} - \left( 1 - \frac{\NNZ}{\TotalParams} \right) \: }
\end{equation}

Note that to compute $\NNZ$, it is necessary to sample from the gating parameters. However, instead of using $\bern(\cdot)$ as the sampling function, we perform a ``maximum-likelihood (ML) draw'' \cite{srinivas2017training} to sample from $\sigmoid( G )$ in order to ensure determinism when calculating the sparsity. ML draw involves thresholding each value $g$ at $0.5$, \ie{} $\sample(\cdot) = \round(\cdot)$. For a model with $R$ layers and gating variables $\phi = \set{G_{1:R}}$, this $\NNZ$ computation takes the following form:

\begin{equation}\label{eq: Pruning: nnz}
    \NNZ = \norm{ \round\left(\sigmoid\left(\phi\right)\right) }^1
\end{equation}

As both the sampling functions $\bern(\cdot)$ and $\round(\cdot)$ are non-differentiable, gradient back-prop has to be performed via an estimator. On this front, \cite{bengio2013estimating} had explored several gradient estimators for stochastic neurons, and the \emph{straight-through estimator} (STE) is found to be simple yet performant. Hence, back-prop is calculated by treating both sampling functions as identity functions, such that the gradients are estimated as $\delta \sample\left(\sigmoid\left(g\right)\right) / \delta \sigmoid\left(g\right) = 1$ such that:

\begin{equation}\label{eq: Pruning: STE}
   \frac{\delta L}{\delta \sigmoid\left(g\right)} = \frac{\delta L}{\delta \sample\left(\sigmoid(g\right))}
\end{equation}

Finally given an image $I$ and caption $C$, the overall cost function for training the image captioning model $\theta$ with gating variables $\phi$ is a weighted combination of captioning loss $L_{c}$ and sparsity loss $L_{s}$:

\begin{equation}\label{eq: Pruning: final loss}
    L \left(\, I,\, C,\, s_{target} \,\right) = L_{c} + \lambda_{s} L_{s}
\end{equation}

Intuitively, it can be seen that the captioning loss term $L_{c}$ is providing a supervised way to learn the saliency of each parameter where important parameters are retained with higher probability whereas unimportant ones are dropped more frequently. On the other hand, the sparsity regularisation term $L_{s}$ pushes down the average value of the gating parameters so that most of them have a value of less than $0.5$ after sigmoid activation. The hyperparameter $\lambda_{s}$ determines the weightage of $L_{s}$. If $\lambda_{s}$ is set too low, the target sparsity level might not be attained (see Sec.~\ref{subsec: Pruning: Experiments: Ablations}). Visualisations of the training progression are given in Sec.~\ref{subsec: Pruning: Experiments: Visualisations}.

\subsection{Inference}
\label{subsec: Pruning: Proposed: Inference}

After model training is completed, all the weight matrices $W$ are transformed into sparse matrices via element-wise multiplication with $G^{b}$. This can be done by sampling from $G$ using $\round(\cdot)$, after which $G$ can be discarded. In other words, the final weights $W^{f}$ are calculated as:

\begin{equation}\label{eq: Pruning: weight masking final}
    W^{f} = W \odot \round\left( \sigmoid\left( G \right) \right)
\end{equation}

The final sparse network can then be stored in appropriate sparse matrix formats such as Coordinate List (COO) or Compressed Sparse Row (CSR) in order to realise compression in terms of storage size. This can be done easily using PyTorch, as it supports parameter saving in COO format (as of release 1.6). Alternatively, regular compression algorithms such as \textit{gzip} can be used to compress the model.

\begin{algorithm}
    \setstretch{1.5}
    \caption{Supermask Pruning (SMP)}
    \label{algo: Pruning: Supermask Pruning}
    \SetKwInput{KwReq}{Require}
    \SetKwComment{Comment}{$\Rightarrow$ }{}
    \SetCommentSty{}    
    \KwReq{Model parameters $\theta = \set{W_{1:R}, B_{1:R}}$, gating parameters $\phi = \set{G_{1:R}}$, sparsity target $s_{target}$, maximum training step $n_{max}$, optimizer $\eta$, training data $D$, loss function $L$}
    \KwOut{Final model parameters $\theta^f = \set{W^f, B^f}$}
    
    $\theta \gets $ ModelSpecificInitializer
    \Comment*{Initialise model parameters}
    $\phi \gets m$
    \Comment*{Initialise gating parameters with constant $m$}
    
    \For{$n \in \set{0, 1, \ldots, n_{max}}$}{
        $(I_n, C_n) \sim D$ 
        \Comment*{Sample a mini-batch of training data}
        $W^{\prime} \gets W \odot \bern\left( \sigmoid\left( G \right) \right)$ 
        \Comment*{Sample effective weights, refer Eq.~\eqref{eq: Pruning: weight masking bern}}
        $\alpha \gets 1 - \frac{1}{2} \left( 1 + \cos \left( \frac{n \pi}{n_{max}} \right) \right)$
        \Comment*{Sparsity loss annealing, refer Eq.~\eqref{eq: Pruning: sparsity anneal}}
        $\NNZ \gets \norm{ \round\left(\sigmoid\left(\phi\right)\right) }^1$ 
        \Comment*{Compute NNZ, refer Eq.~\eqref{eq: Pruning: nnz}}
        $L_{s} \gets \alpha \: \abs*{ \: s_{target} - \left( 1 - \frac{\NNZ}{\TotalParams} \right) \: }$
        \Comment*{Sparsity loss, refer Eq.~\eqref{eq: Pruning: sparsity loss}}
        $L \gets L_{c}\left(I_n, C_n ; W^{\prime}, B \right) + \lambda_{s} L_{s}$
        \Comment*{Final loss, refer Eq.~\eqref{eq: Pruning: final loss}}
        
        $\theta \gets \theta - \eta( \frac{\partial}{\partial \theta} (L) )$ 
        \Comment*{Update model parameters}
        $\phi \gets \phi - \eta( \frac{\partial}{\partial \phi} (L) )$ 
        \Comment*{Update gating parameters}
    }
    
    $W^f \gets W \odot \round( \sigmoid( G ) )$
    \Comment*{Compute final weights, refer Eq.~\eqref{eq: Pruning: weight masking final}}
    $B^f \gets B$ \;
    Discard $\phi$ 
    \Comment*{Gating parameters can be discarded}
\end{algorithm}


\section{Experiments}
\label{sec: Pruning: Experiments}

In this section, we first present the setup of our experiments, followed by the results obtained from over 6,000 GPU hours using 2 Titan X GPUs.

\subsection{Experiment Setup}
\label{subsec: Pruning: Experiments: Setup}

\bft{\itt{Architectures:}} Three different popular image captioning architectures are used in this work: \itt{Soft-Attention} (SA) \cite{xu2015show}, \itt{Up-Down} (UD) \cite{anderson2018bottom} and \itt{Object Relation Transformer} (ORT) \cite{herdade2019image}. SA consists of Inception-V1 \cite{ioffe2015batch}, and a single layer LSTM or GRU with single-head attention function. Other details such as context size, attention size and image augmentation follow \cite{tan2019comic}.
For UD and ORT, we reuse the public implementations\footnote{\url{https://github.com/ruotianluo/self-critical.pytorch/tree/3.2}}\footnote{\url{https://github.com/yahoo/object_relation_transformer}}.

\bft{\itt{Hyperparameters:}} For all training, we utilise Adam \cite{kingma2014adam} as the optimiser, with an epsilon of $1 \times 10^{-2}$ for SA and UD. The SA models were trained for 30 epochs, whereas UD and ORT models were trained for 15 epochs. Cosine LR schedule was used for SA and UD, whereas ORT follows \cite{herdade2019image}. Following \cite{narang2017exploring} and \cite{han2015learning}, lower dropout rates are used for sparse networks to account for their reduced capacity. The rest follows \cite{tan2019comic}.

For Supermask Pruning (SMP), training of the gating variables $\phi$ is done with a higher constant learning rate (LR) of 100 without annealing. This requirement of a higher LR is also noted in \cite{zhou2019deconstructing}. All $\phi$ are initialised to a constant $m = 5.0$.

The other pruning methods are trained as follows. 
\itt{Hard:} Pruning is applied after decoder training is completed. It is then retrained for 10 epochs.
\itt{Gradual:} Pruning begins after the first epoch is completed and ends at half of the total epochs, following the heuristics outlined in \cite{narang2017exploring}. Pruning frequency is 1000. We use the standard scheme where each layer is uniformly pruned. 
\itt{SNIP:} Pruning is done at initialisation using one batch of data. Implementation is based on the authors' code\footnote{\url{https://github.com/namhoonlee/snip-public}}.
\itt{Lottery Ticket:} Winning tickets are produced using hard-blind, hard-uniform and gradual pruning. For a fair comparison with other single-shot pruning methods, we follow the one-shot protocol instead of the iterative protocol.

Inference is performed using beam search without length normalisation.


\bft{\itt{Datasets:}} Experiments are performed on MS-COCO \cite{lin2014microsoft} which is a public English captioning dataset. 
Following prior captioning works, we utilise the ``Karpathy'' split \cite{karpathy2015deep}, which assigns 5,000 images for validation, 5,000 for testing and the rest for training. 
Pre-processing of captions is done following \cite{tan2019comic}.

\bft{\itt{Metrics:}} Evaluation scores are obtained using the publicly available MS-COCO evaluation toolkit\footnote{\url{https://github.com/salaniz/pycocoevalcap}}, which computes BLEU, METEOR, ROUGE-L, CIDEr and SPICE (B, M, R, C, S).

\subsection{Pruning Image Captioning Models}
\label{subsec: Pruning: Experiments: Pruning}

\begin{figure*}[tp]
     \centering
    \begin{subfigure}{.495\linewidth}
        \centering
            \gph{1}{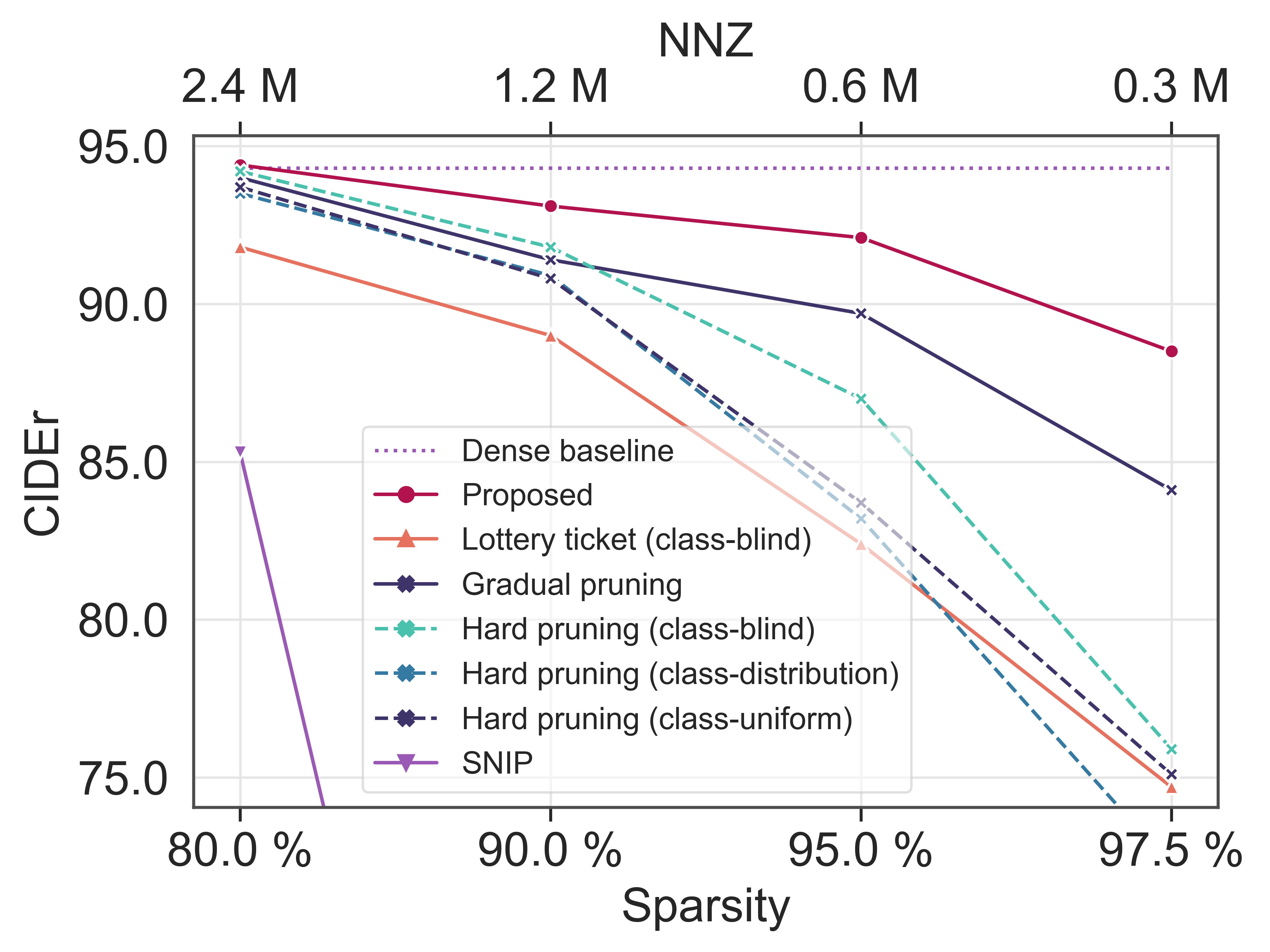}
        \caption{Soft-Attention (SA) LSTM model.}
        \label{fig: Pruning: soft-LSTM-CIDEr-COCO}
    \end{subfigure}
    \begin{subfigure}{.495\linewidth}
        \centering
            \gph{1}{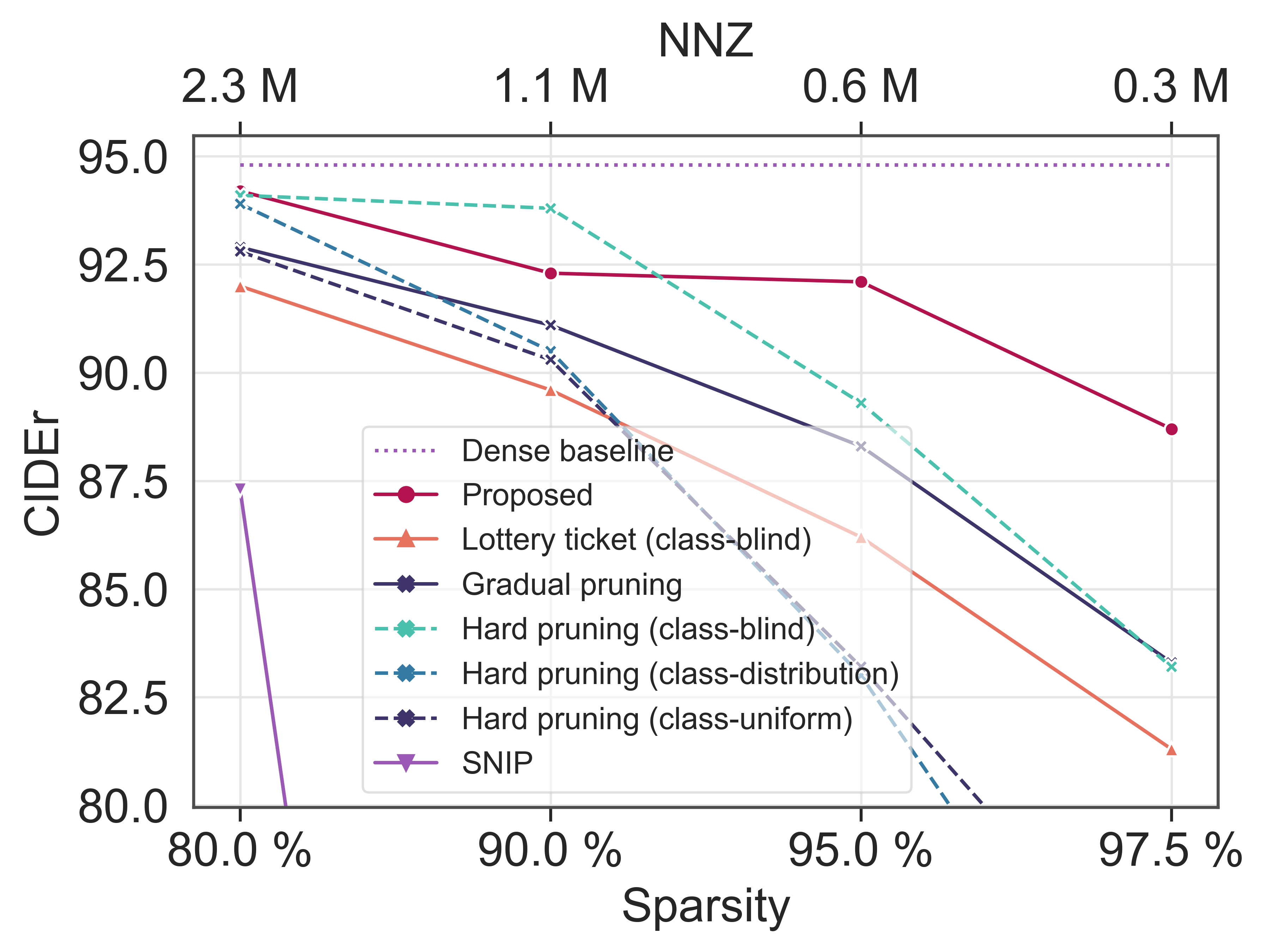}
        \caption{Soft-Attention (SA) GRU model.}
        \label{fig: Pruning: soft-GRU-CIDEr-COCO}
    \end{subfigure}
    \caption{Pruning performance on MS-COCO. $\lambda_{s}$ is set following $\lambda_{s} = \max(5, 0.5 / (1-s_{target}))$. Sparsity, compression and NNZ figures exclude normalisation and bias parameters.}
    \label{fig: Pruning: Soft-Attention COCO}
\end{figure*}

\begin{figure*}[tp]
    \centering
    \begin{subfigure}{.495\linewidth}
        \centering
            \gph{1}{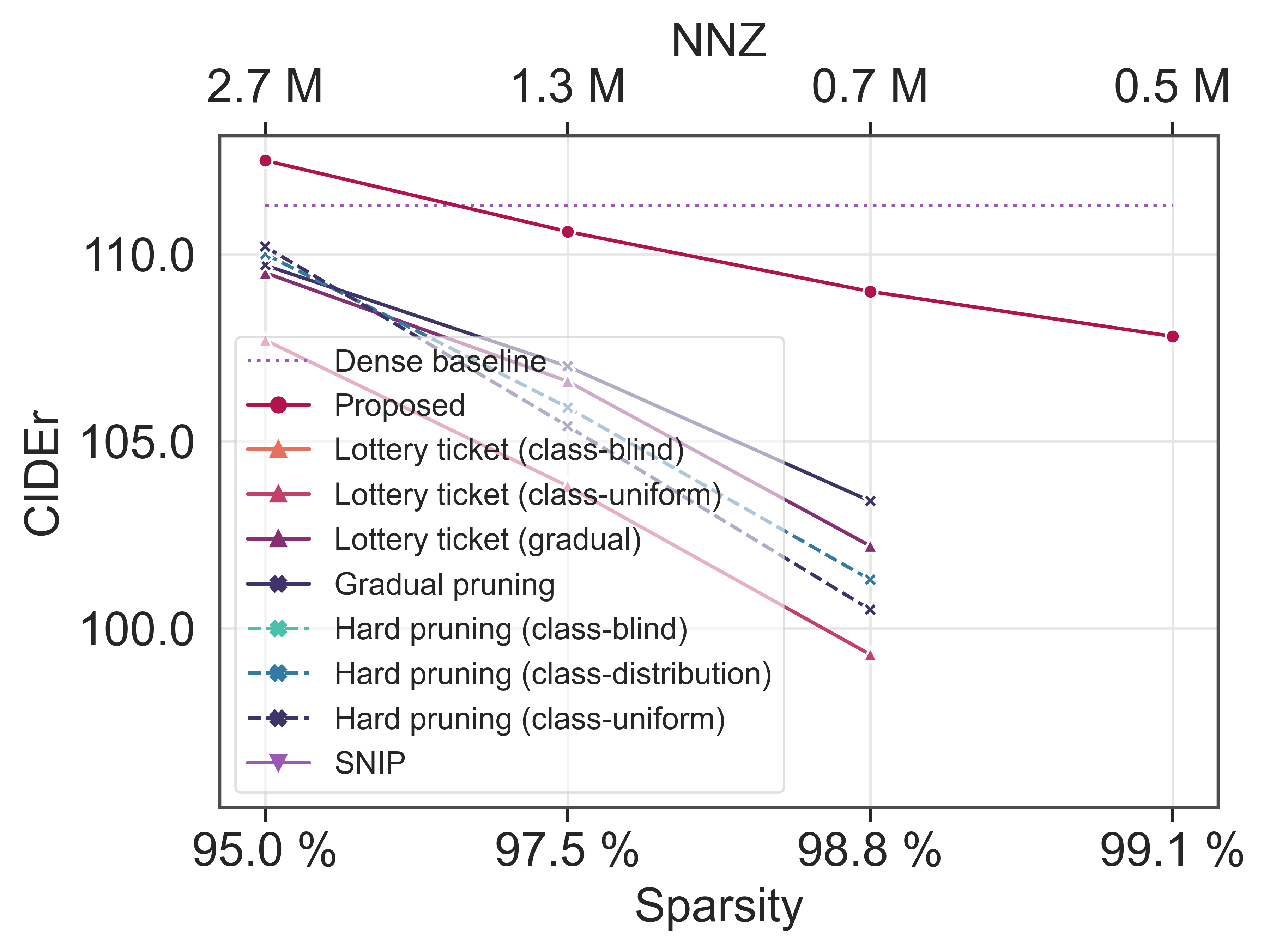}
        \caption{Up-Down (UD) model.}
        \label{fig: Pruning: UpDown-CIDEr-COCO}
    \end{subfigure}
    \begin{subfigure}{.495\linewidth}
        \centering
            \gph{1}{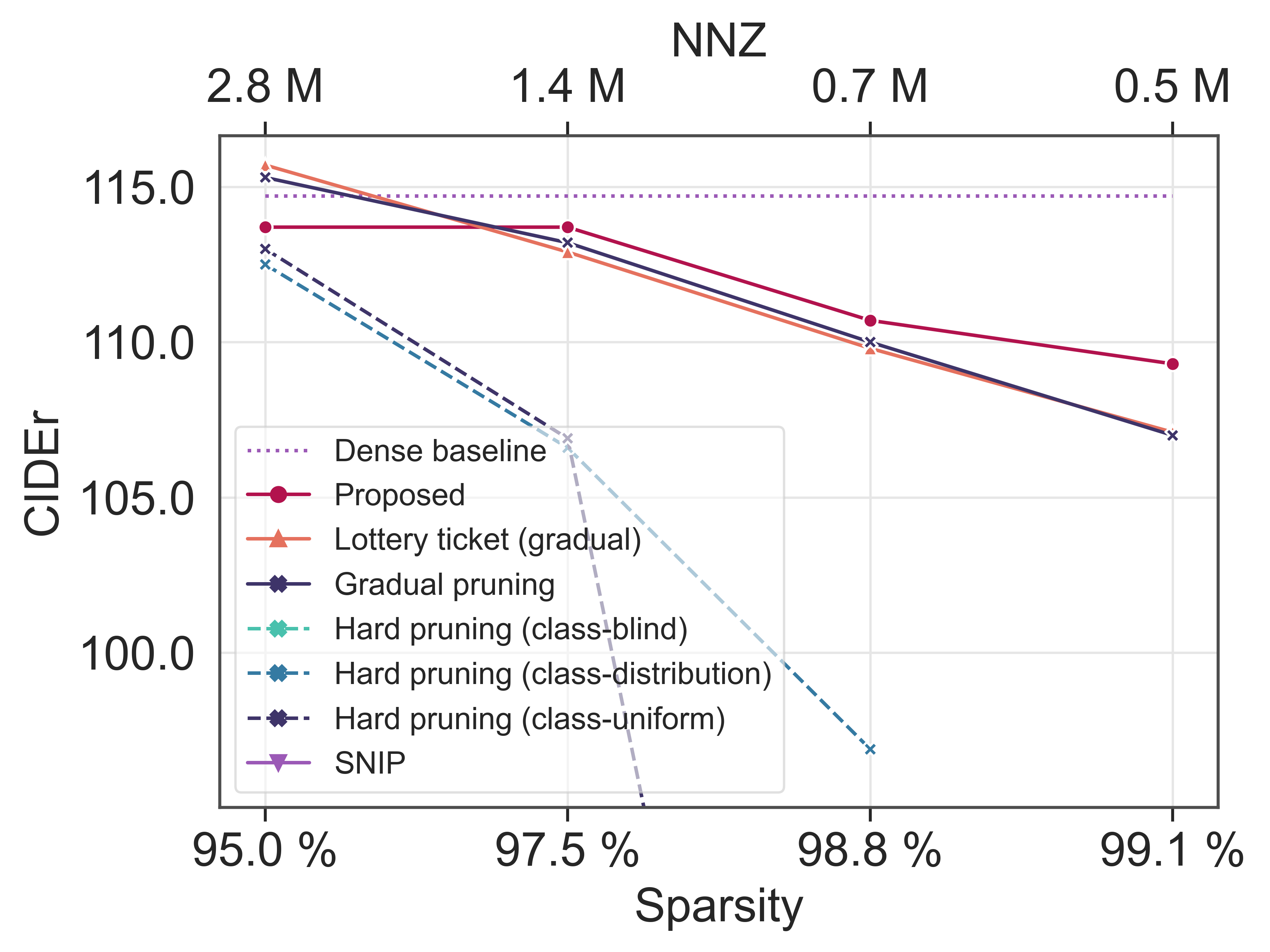}
        \caption{Object Relation Transformer (ORT) model.}
        \label{fig: Pruning: ORT-CIDEr-COCO}
    \end{subfigure}
    \caption{Pruning performance on MS-COCO. Both SNIP and hard-blind methods failed to converge well. In order from 80\% to 99.1\% sparsity, $\lambda_{s}$ is set to $80, 80, 80, 120$ for UD and $120, 120, 80, 120$ for ORT. Sparsity, compression and NNZ figures exclude normalisation and bias parameters.}
    \label{fig: Pruning: UpDown ORT COCO}
\end{figure*}

In this section, we attempt to answer Questions (1) and (2) in Sec.~\ref{sec: Pruning: Introduction} via extensive performance comparisons of the pruning methods at multiple sparsity levels. We first present the pruning results on SA in Fig.~\ref{fig: Pruning: Soft-Attention COCO}, followed by UD and ORT in Fig.~\ref{fig: Pruning: UpDown ORT COCO}. Pruning is applied to all learnable parameters except for normalisation layers and biases. All the results herein were obtained using teacher-forcing with cross-entropy loss.

\itt{Which pruning method produces the best results?} Our proposed end-to-end Supermask Pruning (SMP) method provides a good performance relative to the dense baselines. This observation is valid even at high pruning ratios of 95\% and above. In particular, the relative drops in CIDEr scores for UD and ORT are only marginal ($-3.1\%$ to $-4.7\%$) even at a $111\times$ pruning rate. This is in contrast with competing methods whose performance drops are either double or even triple compared to ours, especially on SA and UD.
To further support this observation, we compute the uniqueness and length of captions produced by our sparse SMP models. Results in Table~\ref{table: Pruning: Uniqueness on MS-COCO} shows that they are largely unaffected by the pruning rate.

Among the competing methods, gradual pruning generally outperforms hard pruning, especially at higher sparsity levels when NNZ falls to 0.6~M and below. On the other hand, the results of LTs indicates that model resetting in a one-shot scenario does not outperform direct application of the underlying pruning method. We note that better results have been reported using iterative prune-reset-train cycles, however that would lead to excessively long training times and unfair comparisons with other pruning methods.

Another notable result is the relatively poor performance of SNIP when applied to image captioning. We can observe in Fig.~\ref{fig: Pruning: Soft-Attention COCO} that the performance of SNIP is acceptable at 80\% sparsity only. Any higher sparsity levels than this quickly led to a collapse in caption quality, as indicated by the metric scores. We tried accumulating the saliency criterion across 100 batches in an attempt to improve the result, but the improvement is limited with a huge gap from the baseline\footnote{To ensure there are no critical errors in our implementation, we had successfully reproduced the results for \itt{LSTM-b} on MNIST with a lower error rate of $1.281\%$ averaged across 20 runs.}. All in all, these results reflect the difficulty of pruning generative models, as well as the importance of testing on larger datasets.

\begin{table}[t]
    \caption{Caption statistics on MS-COCO test set.}
    \label{table: Pruning: Uniqueness on MS-COCO}
    \centering
    \begin{adjustbox}{max width=0.7\linewidth}
    \begin{tabular}{ `l ~c ~c ~c }
        \toprule
        \fmtr{2.35}{Approaches} & \fmtr{2.35}{\makecell{Sparsity\\(\%)}} & \multicolumn{2}{c}{Caption stats.} \\
                                                                            \cmidrule(lr){3-4}
        \null                   & \null        & Unique (\%)  & Av. len. \\
        
        \midrule
        SA LSTM Dense           & -            & 42.1 & 9.09 \\
        SMP                     & 80.0         & 42.5 & 9.11 \\
        \null                   & 97.5         & 44.4 & 8.99 \\
        \midrule
        SA GRU Dense            & -            & 42.4 & 9.15 \\
        SMP                     & 80.0         & 43.1 & 9.13 \\
        \null                   & 97.5         & 42.0 & 8.94 \\
        \midrule
        UD Dense                & -            & 53.0 & 9.46 \\
        SMP                     & 95.0         & 58.6 & 9.46 \\
        \null                   & 99.1         & 61.7 & 9.30 \\
        \midrule
        ORT Dense               & -            & 61.2 & 9.52 \\
        SMP                     & 95.0         & 62.4 & 9.46 \\
        \null                   & 99.1         & 61.1 & 9.28 \\
        
        \bottomrule
    \end{tabular}
    \end{adjustbox}
\end{table}

\begin{table*}[ht]
    \caption{Single-model comparison with captioning SOTA. NNZ and model size calculations exclude CNN.}
    \label{table: Pruning: MS-COCO SOTA}
    \centering
    \begin{adjustbox}{max width=\linewidth}
    \begin{threeparttable}
    \begin{tabular}{ `l  ~c ~c ~c ~c ~c ~c ~c }
        \toprule
        \fmtr{2.35}{Approaches} & \fmtr{2.35}{NNZ (M)} & \fmtr{2.35}{\makecell{Model size\\(MB)}} & \multicolumn{5}{c}{MS-COCO test scores} \\
                                                                                                    \cmidrule(lr){4-8}
        \null                                   & \null     & \null         & B-1  & B-4  & M     & C     & S    \\
        
        \midrule
        DeepVS \cite{karpathy2015deep}          & -         & -             & 62.5 & 23.0 & 19.5  & ~66.0 & -    \\
        SA \cite{xu2015show}            & 11.9 \tnote{a}  & ~70.1 \tnote{a} & 70.7 & 24.3 & 23.9  & -     & -    \\
        ALT-ALTM \cite{ye2018attentive}         & -         & -             & 75.1 & 35.5 & 27.4  & 110.7 & 20.3 \\
        ARL \cite{wang2020learning}             & -         & -             & 75.9 & 35.8 & 27.8  & 111.3 & -    \\
        Att2all \cite{rennie2017self}   & 46.3 \tnote{a}  & 185.3 \tnote{a} & -    & 34.2 & 26.7  & 114.0 & -    \\ 
        UD \cite{anderson2018bottom}    & 53.2 \tnote{a}  & 212.6 \tnote{a} & 79.8 & 36.3 & 27.7  & 120.1 & 21.4 \\
        ORT \cite{herdade2019image}     & 55.4 \tnote{a}  & 232.2 \tnote{a} & 80.5 & 38.6 & 28.7  & 128.3 & 22.6 \\
        M2 \cite{cornia2020meshed}              & -         & -             & 80.8 & 39.1 & 29.2  & 131.2 & 22.6 \\
        
        \midrule
        UD (95.0\%)                    \rbf & 2.7   &  53.3                 & 79.7 & 38.5 & 27.9  & 124.9 & 20.9 \\ 
          (99.1\%, float16)                 & 0.5   & ~~\bft{8.7} \tnote{b} & 78.9 & 37.2 & 27.3  & 120.1 & 20.0 \\
        
        \midrule
        ORT (95.0\%)                  \rbf & 2.9    &  66.2                 & 80.5 & 39.1 & 28.5  & 129.4 & 21.6 \\ 
          (99.1\%, float16)                & 0.6    & ~\bft{14.5} \tnote{b} & 79.4 & 37.6 & 27.8  & 124.3 & 20.9 \\
        
        \bottomrule
    \end{tabular}
    \begin{tablenotes}
        \item[a] Based on reimplementation, size in float32.
        \item[b] Size in float32: 9.7 MB (UD), 20.8 MB (ORT)
    \end{tablenotes}
    \end{threeparttable}
    \end{adjustbox}
\end{table*}

\begin{figure*}[ht]
    \centering
    \begin{subfigure}{.495\linewidth}
        \centering
            \gph{1}{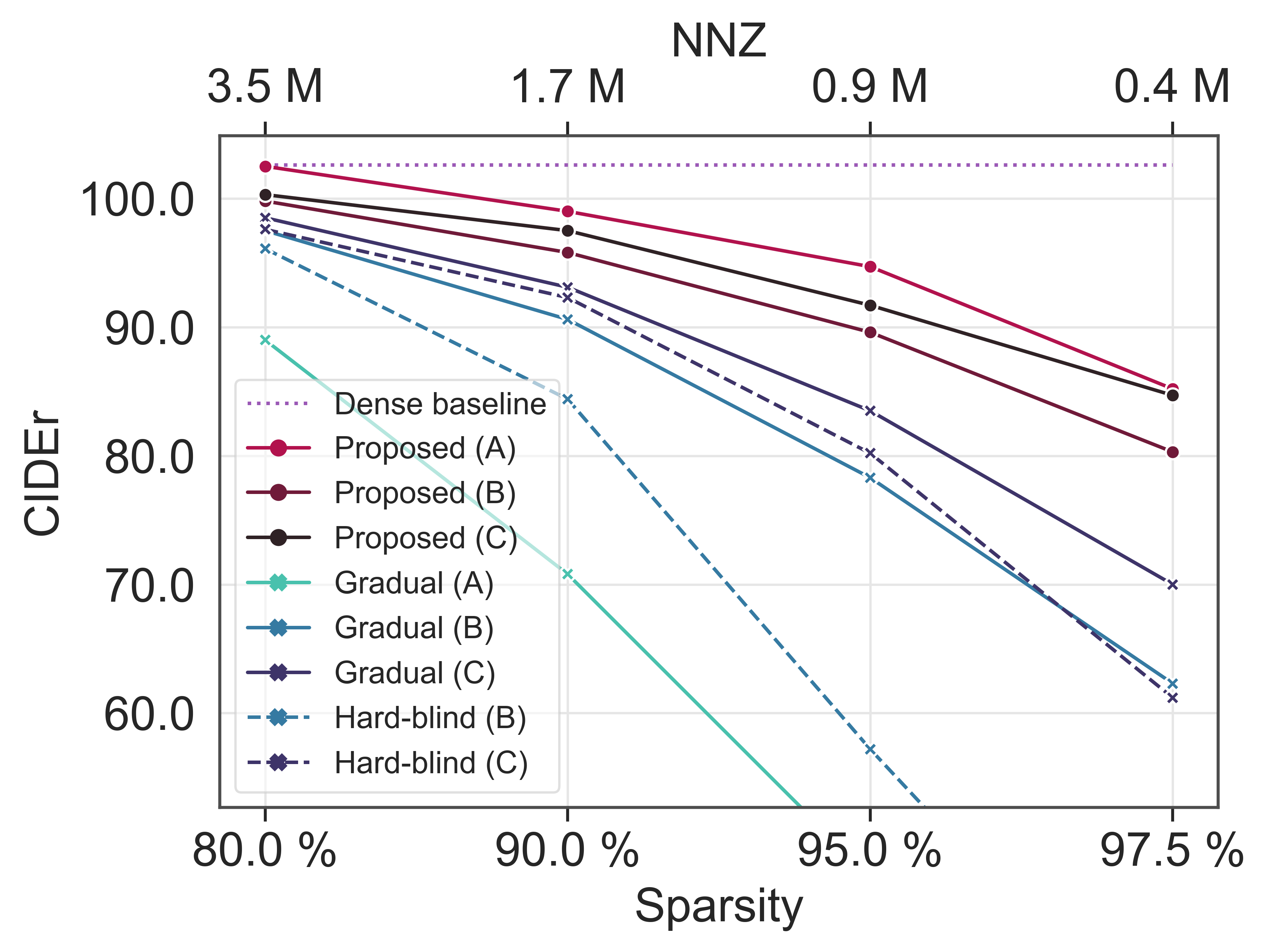}
        \caption{Soft-Attention (SA) LSTM model.}
        \label{fig: Pruning: Sparse Inception-V1 + LSTM CIDEr}
    \end{subfigure}
    \begin{subfigure}{.495\linewidth}
        \centering
            \gph{1}{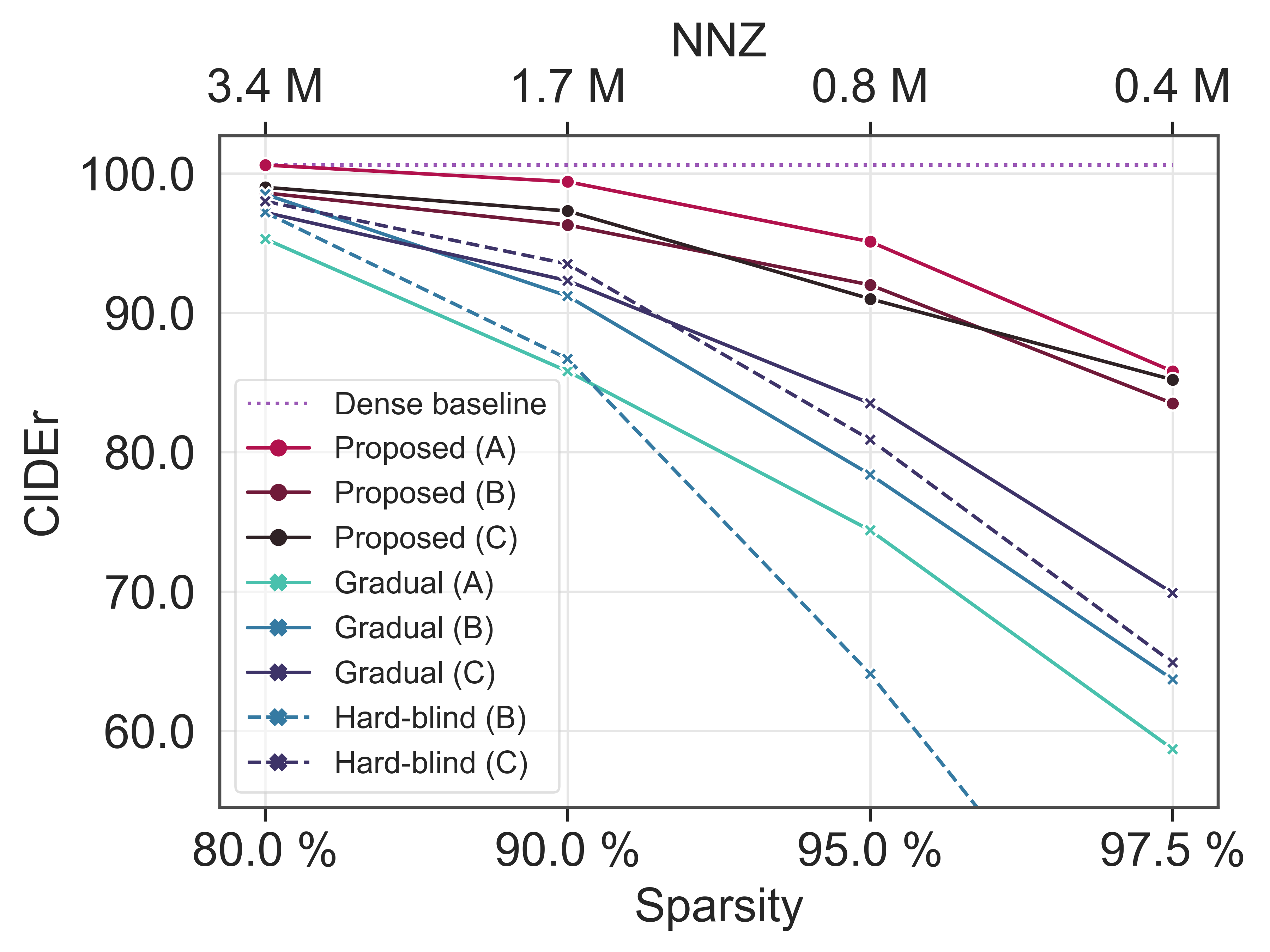}
        \caption{Soft-Attention (SA) GRU model.}
        \label{fig: Pruning: Sparse Inception-V1 + GRU CIDEr}
    \end{subfigure}
    \caption{Pruning performance on MS-COCO when both encoder and decoder are pruned. Sparsity, compression and NNZ figures exclude normalisation and bias parameters.}
    \label{fig: Pruning: Fine-tune MSCOCO}
\end{figure*}

\itt{Is there an ideal sparsity?} A broad trend that emerged from Fig.~\ref{fig: Pruning: Soft-Attention COCO} and \ref{fig: Pruning: UpDown ORT COCO} is that the model performance is more dependent on the remaining NNZ parameters after pruning, rather than the sparsity level. Both the UD and ORT models, which are about $4\times$ larger than the SA model, can achieve substantially higher sparsity. On the extreme end, we were able to prune 99.1\% of parameters from the networks, while suffering only $-3.5$ CIDEr points for UD and $-5.4$ CIDEr points for ORT.

In addition, there are indeed ideal sparsity levels where sparse models can either match or outperform their dense counterparts. This occurs at an 80\% sparsity for SA, and at a 95\% sparsity for both UD and ORT. We did not further investigate the performance of these models at lower sparsities, as although it is reasonable to expect better performance, the model sizes also increase substantially.


All in all, these results showcase the strength of SMP across pruning ratios from 80\% to 99.1\%, while managing good performance relative to the dense baselines and other pruning methods.

\begin{table}[ht]
    \caption{Comparison with H-LSTM \cite{dai2020grow,dai2019nest}. Both encoder and decoder of the SMP models are pruned.}
    \label{table: Pruning: H-LSTM}
    \centering
    \begin{adjustbox}{max width=0.95\linewidth}
    \begin{tabular}{ `l  ~c ~c ~c ~c ~c ~c }
        \toprule
        \fmtr{2.35}{Approaches}     & \multicolumn{2}{c}{RNN (K)} & \multicolumn{4}{c}{MS-COCO test scores} \\
                                      \cmidrule(lr){2-3}             \cmidrule(lr){4-7}
        \null               & NNZ   & FLOP  & B-1  & B-4  & C     & S    \\
        
        \midrule
        SMP (80\%)          &  562  & 1128  & 73.9 & 33.4 & 102.5 & 18.8 \\ 
        
        \midrule
        H-LSTM + GP         &  394  & 670   & 71.9 & -    & ~95.4 & -    \\
        SMP (90\%) \rbf     &  270  & 545   & 72.9 & 32.8 & ~99.0 & 18.3 \\ 

        \midrule
        H-LSTM + GP         &  163 & 277    & 71.4 & -    & ~93.3 & -    \\
        SMP (95\%) \rbf     &  116 & 236    & 72.0 & 31.7 & ~94.7 & 17.5 \\ 
        
        \bottomrule
    \end{tabular}
    \end{adjustbox}
\end{table}

\subsection{SOTA Comparison}
\label{subsec: Pruning: Experiments: SOTA}

In this section, we compare models pruned using our proposed SMP against both H-LSTM by Dai \etal{} \cite{dai2020grow,dai2019nest} and standard captioning SOTA approaches.

\textbf{\textit{H-LSTM comparison:}} In Table~\ref{table: Pruning: H-LSTM}, we provide the compression rate and model performance comparisons with \cite{dai2020grow}. Our SMP models are SA models trained and fine-tuned using teacher-forcing. As it can be seen, both of the SMP models at 90\% and 95\% sparsities with smaller RNN sizes outperform H-LSTM on both BLEU-4 and CIDEr. Furthermore, SMP does not require the expensive and time-consuming process of ``grow-prune-retrain'' cycles as required by \cite{dai2019nest}.

\textbf{\textit{SOTA comparison:}} To demonstrate that sparse SMP models are competitive with standard SOTA works, we compare UD and ORT models pruned using SMP against several SOTA approaches in Table~\ref{table: Pruning: MS-COCO SOTA}. We optimised our models for BLEU-4 and CIDEr using SCST \cite{rennie2017self}, but with the mean of rewards as baseline following \cite{cornia2020meshed,luo2020better}. Sparse models are saved in PyTorch COO format. For \itt{float16} models, weights are converted back to single-precision prior to computation.

From the results, it is evident that our pruned models are still capable of obtaining good captioning performance. In fact, our 95\% sparse UD and ORT models managed to outperform their original dense counterparts. This is consistent with the findings in Section~\ref{subsec: Pruning: Experiments: Pruning}, which found that 95\% sparsity is ideal. Finally, despite having a relatively small model size of 10 MB and 21 MB, our 99.1\% sparse models provided good results as well. The 99.1\% sparse UD model, in particular, is able to match the dense UD model on CIDEr while outperforming it on BLEU-4.

\subsection{Pruning Sequence for Encoder}
\label{subsec: Pruning: Experiments: Pruning Sequence}

In this section, we attempt to answer Question (3), which asks: \itt{what is the ideal prune-finetune sequence for the encoder ?} To answer this, we devised three prune-finetune schemes for the SA model as follows:

\bft{Scheme A}: Start from scratch: Train the decoder while pruning both the encoder and decoder. Then, fine-tune both with gating $\phi$ frozen (\ie{} not updated).

\bft{Scheme B}: Start from a trained decoder: Fine-tune and prune both the encoder and decoder.

\bft{Scheme C}: Start from a trained and pruned decoder: Fine-tune both the encoder and decoder, but only prune the encoder. Decoder $\phi$ are left frozen.

We paired each of the schemes with three pruning methods from the previous section, namely i) class-blind hard pruning, ii) gradual pruning and iii) SMP. All learnable parameters were pruned except for normalisation layers and biases. For schemes where gating parameters $\phi$ are frozen, we still apply $\bern(\cdot)$ to sample from $\sigmoid(\phi)$. However, we also found that there is minimal difference in the final performance when $\round(\cdot)$ is used instead. Scheme A is not evaluated for hard-blind as it requires a trained model prior to pruning. 

From Fig.~\ref{fig: Pruning: Fine-tune MSCOCO}, it is evident that Scheme A produces polarised results. Specifically, it is the best when paired with SMP, yet is the worst with gradual and hard-blind. On the other hand, Scheme C is consistently favoured over Scheme B for all three pruning methods. This shows that better performance can be attained when pruning and training for the decoder are done in-parallel rather than separately.

Comparing the three different pruning methods, we can see that the trends are consistent with the results obtained for decoder pruning in the previous section. Across different sparsity levels, our SMP method produces the best performance. At 80\% sparsity, there is barely any performance loss relative to the baselines, with a mere $-0.1$ in CIDEr score for LSTM and no difference for GRU. At the other extreme with \emph{2.5\% of parameters}, we managed CIDEr scores of $85.2$ for LSTM and $85.8$ for GRU, while gradual and hard-blind scored $70.0$ and below.

\subsection{Large-Sparse vs Small-Dense}
\label{subsec: Pruning: Experiments: Large-Sparse}

\begin{table}[t]
    \caption{Large-sparse versus small-dense models. Both encoder and decoder are pruned and fine-tuned.}
    \label{table: Pruning: Sparse vs Dense MobileNet-V1 LSTM}
    \centering
    \begin{adjustbox}{max width=\linewidth}
    \begin{tabular}{ `l  ~c ~c ~c ~c ~c ~c ~c }
        \toprule
        \fmtr{2.35}{Approaches} & \multicolumn{2}{c}{Cost (M)} & \multicolumn{5}{c}{MS-COCO test scores} \\
                               \cmidrule(lr){2-3}                 \cmidrule(lr){4-8}
        \null               & NNZ   & FLOP              & B-1  & B-4  & M    & C    & S    \\ 
        
        \midrule
        Dense-L             & 15.33 & 4376              & 73.2 & 32.6 & 25.0 & 98.0 & 18.1 \\ 
        Sparse (80.0\%)     & \bft{3.08} & \bft{901}    & 72.1 & 31.4 & 24.9 & 95.9 & 17.7 \\ 
        
        \midrule
        Dense-M             & 3.37  & 731               & 70.3 & 29.3 & 23.3 & 86.9 & 16.4 \\ 
        Sparse (90.0\%)     & \rbf 1.56 & 533           & 72.2 & 31.4 & 24.7 & 94.9 & 17.7 \\ 
        
        \midrule
        Dense-S             & 2.67  & 340               & 67.1 & 26.7 & 21.7 & 76.6 & 14.7 \\ 
        Sparse (95.0\%)     & \rbf 0.80 & 307           & 70.0 & 29.2 & 23.3 & 86.0 & 16.1 \\ 
        Sparse (97.5\%)     & \bft{0.42} & \bft{195}    & 66.6 & 25.8 & 21.3 & 73.1 & 14.3 \\ 
       
        \bottomrule
    \end{tabular}
    \end{adjustbox}
\end{table}

\textit{Can a sparse model outperform a smaller dense model?} Towards that end, empirical results are given in Table~\ref{table: Pruning: Sparse vs Dense MobileNet-V1 LSTM}. 
All models are based on SA in Sec.~\ref{subsec: Pruning: Experiments: Pruning} but with different CNN and LSTM sizes. The CNNs are MobileNet-V1: \textit{Dense-L} and \textit{Sparse} have a width multiplier of $1.0$; \textit{Dense-M} and \textit{-S} have a width of $0.5$ and $0.25$ respectively. Moreover, \textit{Dense-M} and \textit{-S} have a word embedding size of $88$, with an attention and LSTM size of $128$. The FLOP counts are for generating a 9-word caption from a 224$\times$224 image using a beam size of $3$ (average caption length is 9). \textit{Sparse} models are pruned using SMP.

Comparing models with similar metric scores, large-sparse models often have smaller NNZ and FLOP counts than their dense counterparts. Notably, a 95\% sparse model can provide comparable performance as \textit{Dense-M} that is larger and heavier ($4.3\times$ NNZ and $2.37\times$ FLOP). This further showcases the strength of model pruning and solidifies the observations made in works on RNN pruning \cite{zhu2017prune,narang2017exploring}. 

\subsection{Qualitative Results and Visualisations}
\label{subsec: Pruning: Experiments: Visualisations}

In this section, we present examples of captions generated, as well as visualisations of training progression, final layer-wise sparsities and weight distribution of sparse SMP models.

\begin{figure*}[t]
    \centering
    \gph{0.82}{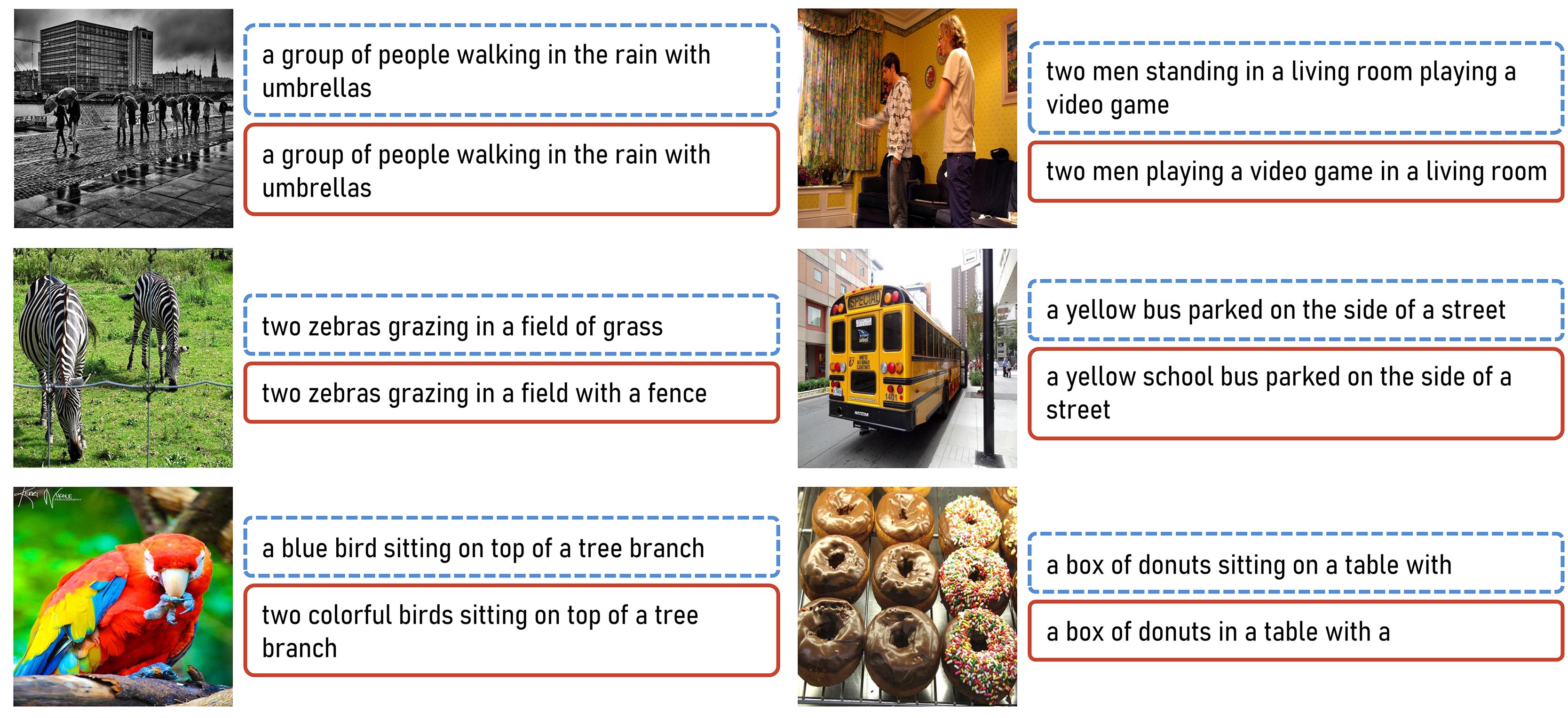}
    \caption{MS-COCO captions generated by 99.1\% sparse UD (dashed blue box) and ORT (solid red box) models.}
    \label{fig: Pruning: Generated captions}
\end{figure*}

\textbf{\textit{Qualitative results:}} Figure~\ref{fig: Pruning: Generated captions} shows the captions produced by our sparse UD and ORT models from Table~\ref{table: Pruning: MS-COCO SOTA}. From the samples, we can see that the overall caption quality is satisfactory with sufficient details, such as umbrellas, living room, fence and school bus. Object counts are largely correct except for 5th image in which a bird is confused for two. The last image shows captions with bad endings, which is a side-effect of SCST optimisation.

\begin{figure}[t]
    \centering
    \gph{0.9}{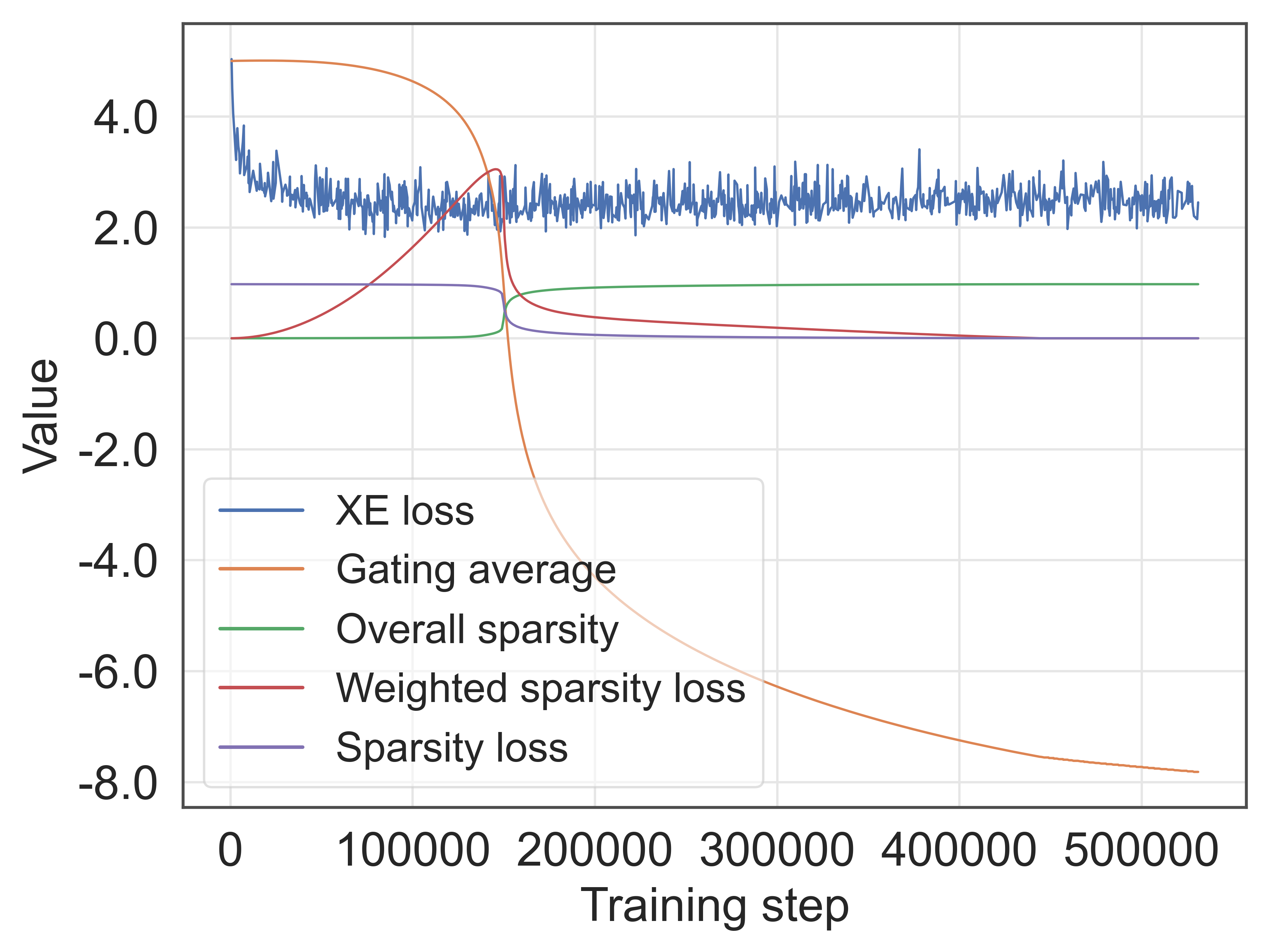}
    \caption{Training progression of the 97.5\% sparse SA model using SMP on MS-COCO. ``Gating average'' is the average value of gating variables $\phi$; ``XE loss'' refers to $L_{c}$ in Eq.~\eqref{eq: Pruning: final loss}; ``Weighted sparsity loss'' refers to $\lambda_{s} L_{s}$ in Eq.~\eqref{eq: Pruning: final loss} ($\lambda_{s} = 20$); ``Sparsity loss'' refers to $L_{s}$ without cosine annealing in Eq.~\eqref{eq: Pruning: sparsity loss}.}
    \label{fig: Pruning: Training progression}
\end{figure}

\textbf{\textit{Training progression:}} Meanwhile in Fig.~\ref{fig: Pruning: Training progression}, we can observe the effects of cosine annealing $\alpha$ from Eq.~\eqref{eq: Pruning: sparsity anneal} and the sparsity regularisation weightage $\lambda_{s}$ from Eq.~\eqref{eq: Pruning: final loss} on the final weighted sparsity loss term. 
Loss annealing allows the model to focus on learning useful representations to solve the captioning task during the early stages of training, and then move towards a sparse solution during the middle to late stages when the training has stabilised. Note that whereas both figures show that sparsity levels only start to increase at around 25\% of total training steps, the pruning process actually began much earlier. The average value of gating variables $\phi$ began to decrease around 10\% into training, and continued to drop towards $-8.0$ throughout later stages of the training process. We can also observe that the training loss (XE loss) remained relatively stable throughout the training and pruning process even for a 97.5\%-sparse model.

\begin{figure*}[t]
    \centering
    \begin{subfigure}{.32\linewidth}
        \centering
            \gph{1}{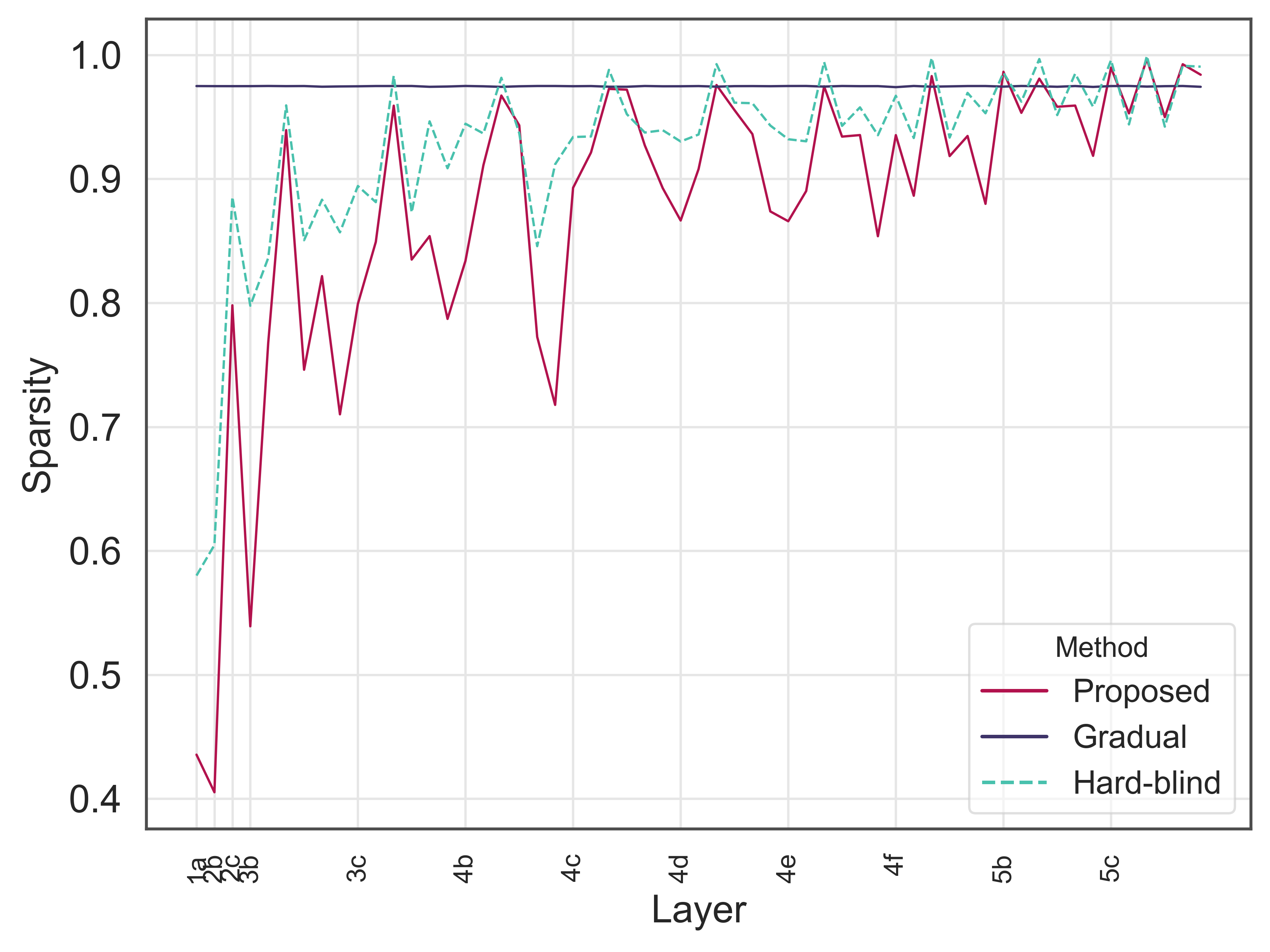}
        \caption{Inception-V1.}
        \label{fig: Pruning: Layerwise sparsity Inception-V1}
    \end{subfigure}
    \begin{subfigure}{.32\linewidth}
        \centering
            \gph{1}{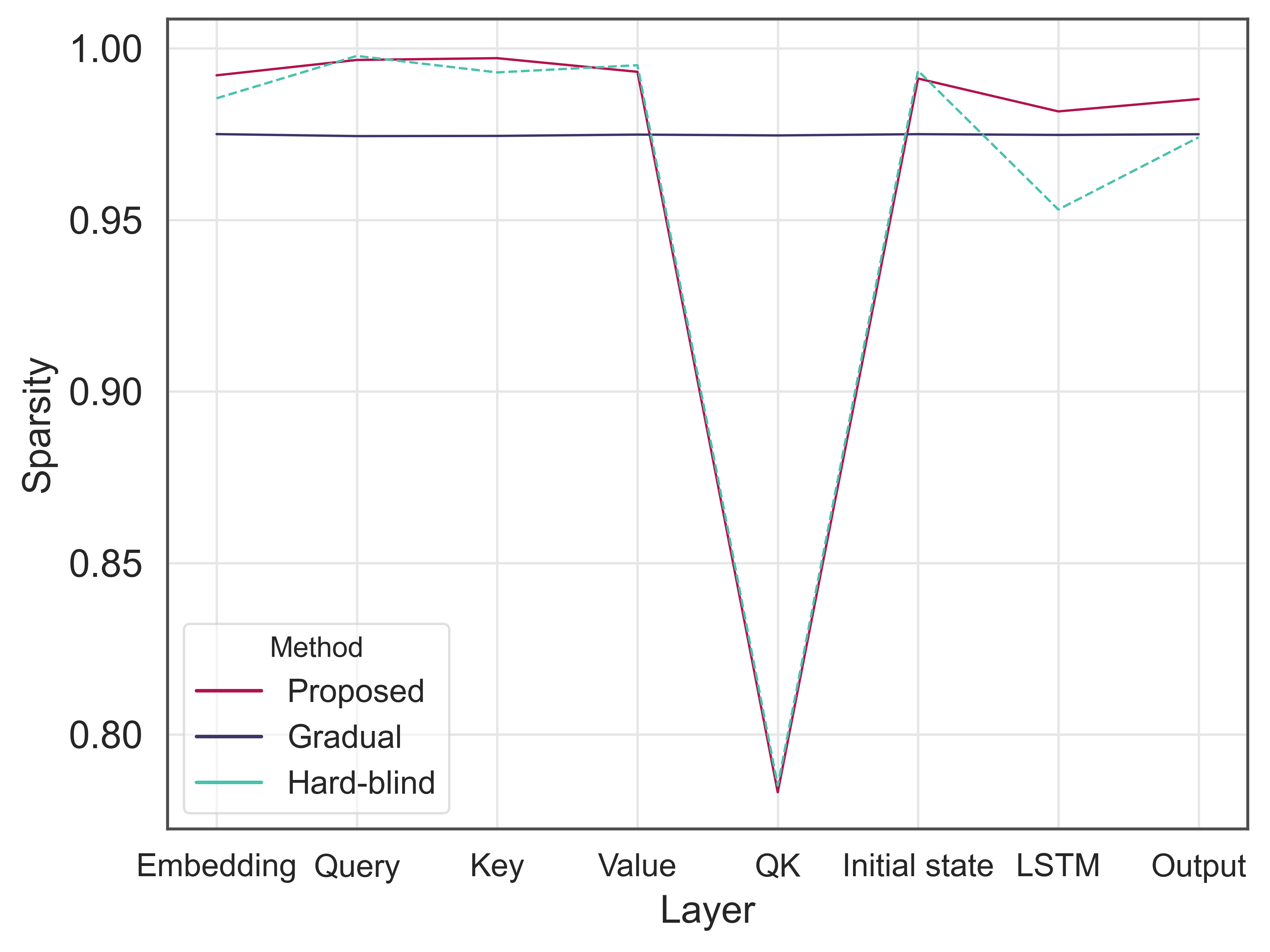}
        \caption{LSTM decoder.}
        \label{fig: Pruning: Layerwise sparsity LSTM decoder}
    \end{subfigure}
    \begin{subfigure}{.32\linewidth}
        \centering
            \gph{1}{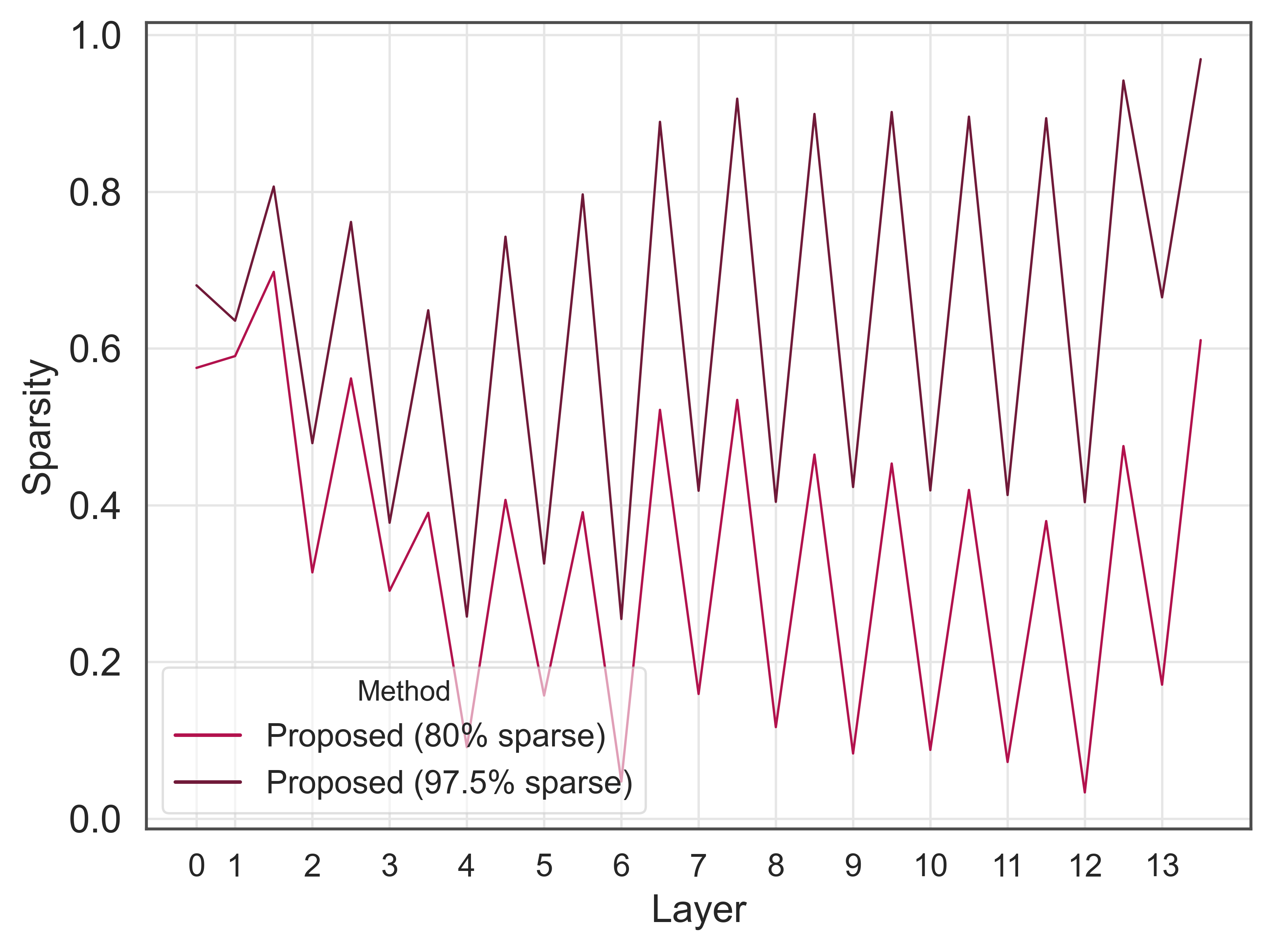}
        \caption{MobileNet-V1.}
        \label{fig: Pruning: Layerwise sparsity MobileNet-V1}
    \end{subfigure}
    \caption{Final layer-wise sparsity levels of SA model. (a) and (b) are 97.5\% sparse.}
    \label{fig: Pruning: Layerwise sparsity}
\end{figure*}

\textbf{\textit{Layer-wise sparsities:}} For Inception-V1 encoder (Fig.~\ref{fig: Pruning: Layerwise sparsity Inception-V1}) pruned using SMP or hard-blind pruning, we can see that earlier convolution layers with fewer parameters are pruned less heavily than later layers. This behaviour is consistent with findings in \cite{elsen2020fast}. We can also see that the $3\times3$ convolution kernel of the second branch of each Inception module is pruned the most compared to the rest.

For LSTM decoder (Fig.~\ref{fig: Pruning: Layerwise sparsity LSTM decoder}), SMP and hard-blind pruning consistently prune ``QK'' layer (the second layer of the 2-layer attention MLP) the least, whereas ``Key'' and ``Query'' layers were pruned most heavily. Finally, ``Embedding'' consistently receives more pruning than ``Output'' despite having fewer parameters. This may indicate that there exists substantial information redundancy in the word embeddings matrix as noted in \cite{shi2018structured,tan2019comic}. 

For MobileNet-V1 encoder (Fig.~\ref{fig: Pruning: Layerwise sparsity MobileNet-V1}), SMP consistently prunes point-wise ($1\times1$) convolution kernels significantly more than depth-wise kernels. This is a desirable outcome as point-wise operations overwhelmingly dominate the computation budget of separable convolutions both in terms of FLOP count and parameters \cite{howard2017mobilenets,elsen2020fast}.

\begin{figure*}[t]
    \centering
    \begin{subfigure}{.325\linewidth}
        \centering
            \gph{1}{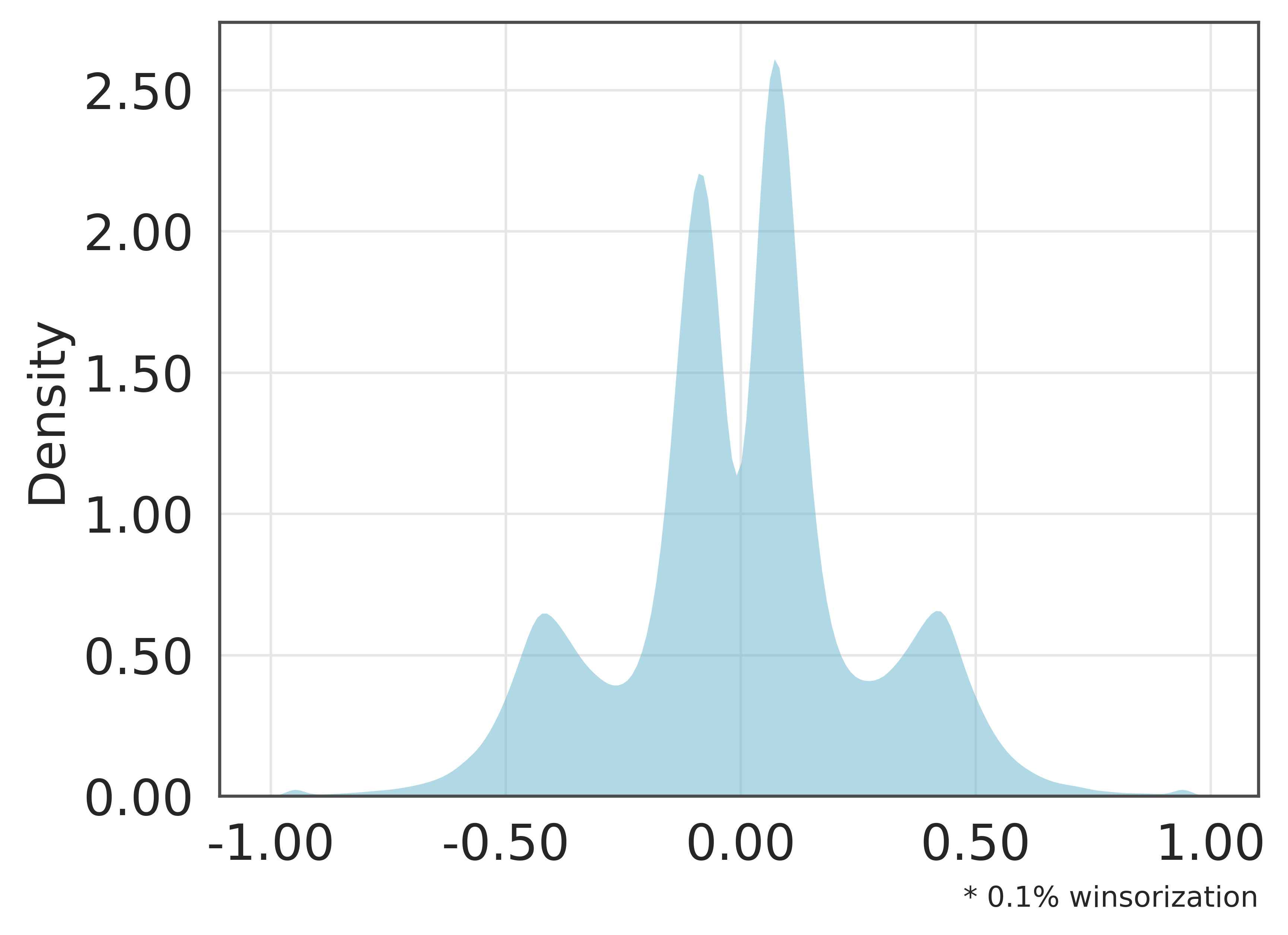}
        \caption{SA model, 97.5\% sparse.}
        \label{fig: Pruning: Dist SA}
    \end{subfigure}
    \begin{subfigure}{.325\linewidth}
        \centering
            \gph{1}{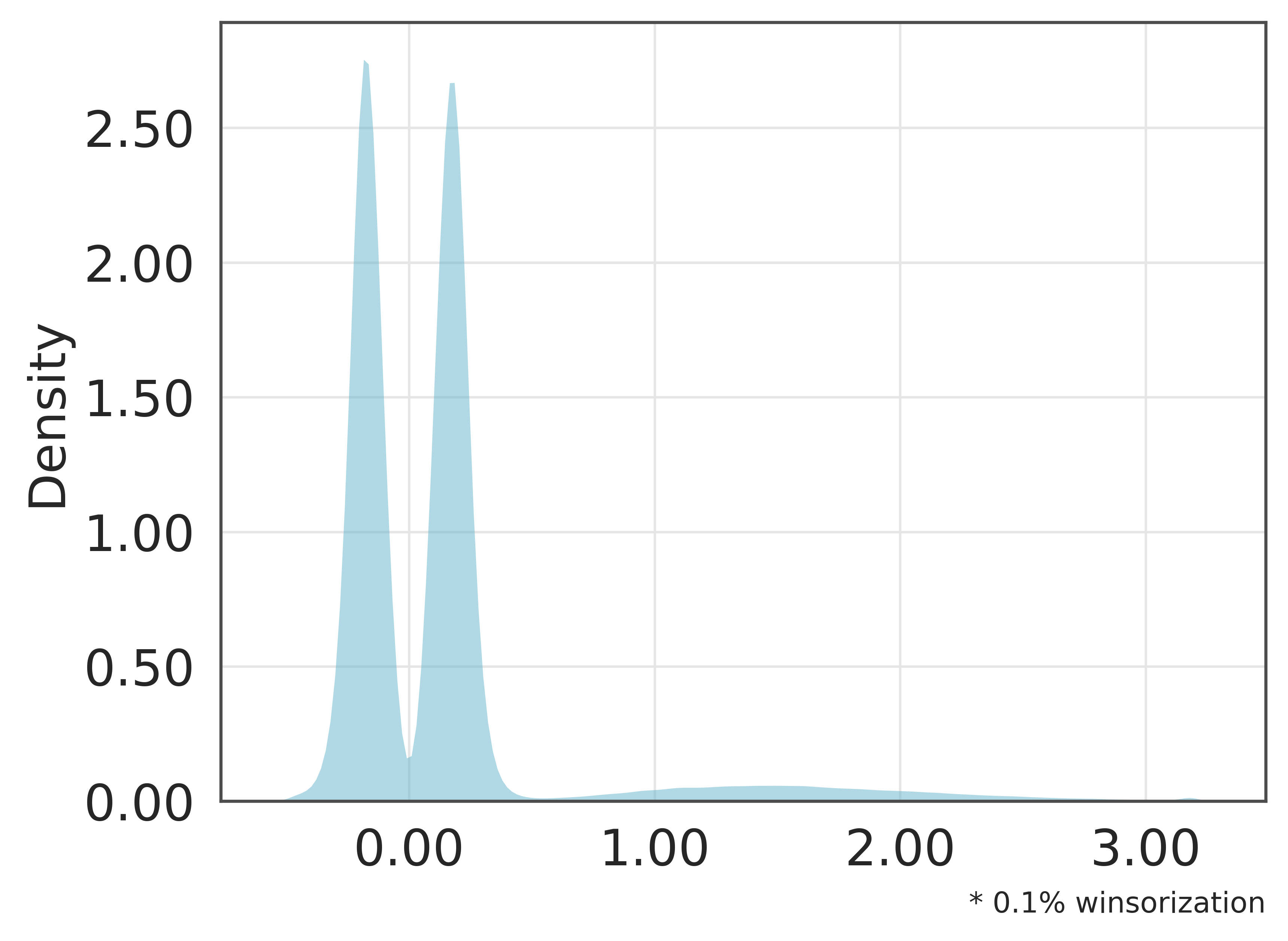}
        \caption{UD model, 99.1\% sparse.}
        \label{fig: Pruning: Dist UD}
    \end{subfigure}
    \begin{subfigure}{.325\linewidth}
        \centering
            \gph{1}{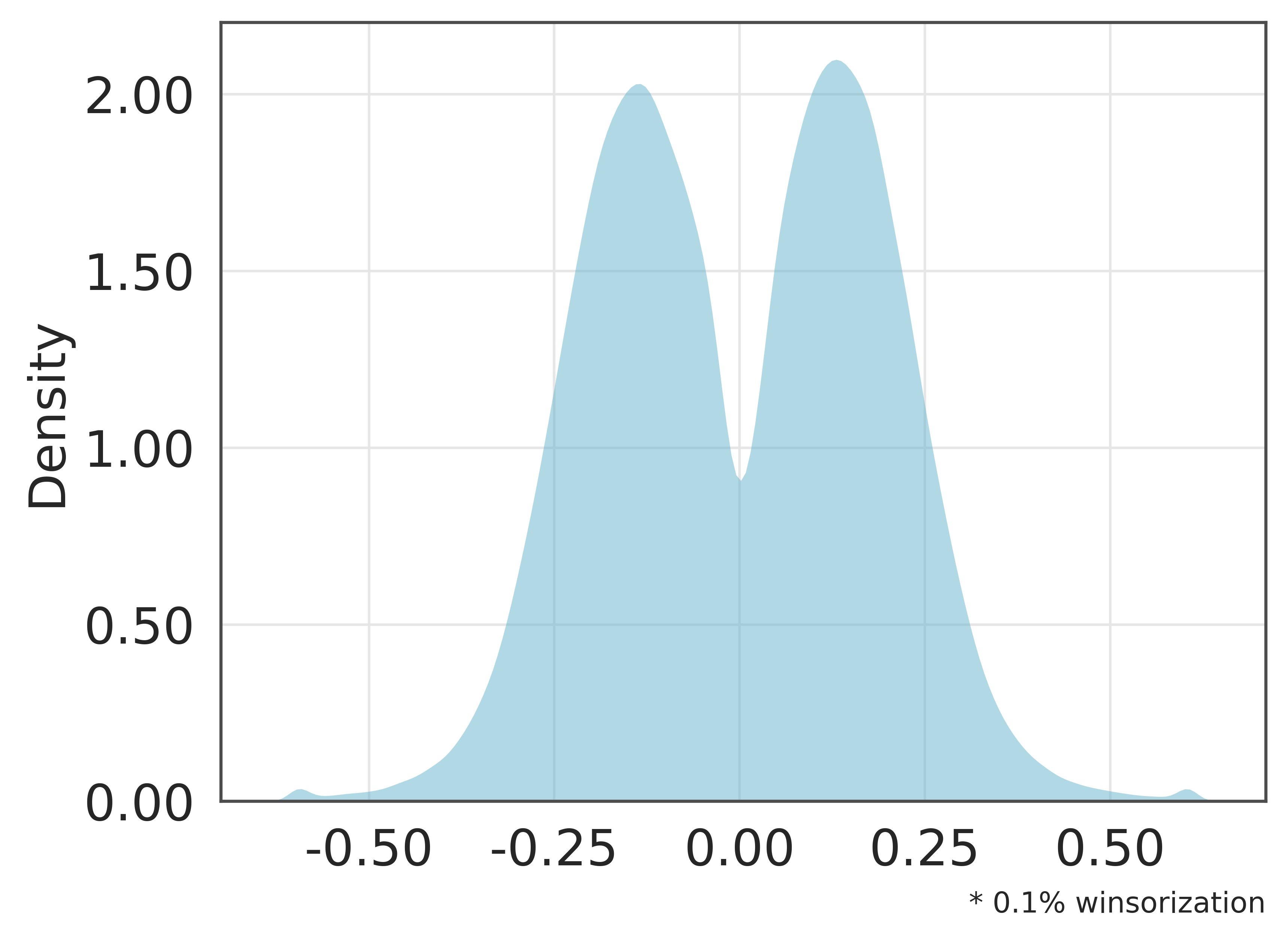}
        \caption{ORT model, 99.1\% sparse.}
        \label{fig: Pruning: Dist ORT}
    \end{subfigure}
    \caption{Distribution of non-zero weights in models pruned using SMP.}
    \label{fig: Pruning: Distribution non-zero}
\end{figure*}

\textbf{\textit{Weight distribution:}} Lastly, distributions of non-zero weights are visualised via kernel density estimation plots in Fig.~\ref{fig: Pruning: Distribution non-zero}. We can see that the remaining weights have a bi-modal distribution centred around zero. Notably, there are still considerable amounts of small-magnitude weights after fine-tuning, even for extremely-sparse models. 

\subsection{Ablation Studies}
\label{subsec: Pruning: Experiments: Ablations}

This section investigates the impact of hyperparameters on the performance of SMP.

\begin{table}[t]
    \caption{Effects of varying initialisation value $m$ of gating parameters $\phi$ on a 80\% sparse SA model.} 
    \label{table: Pruning: Ablation: Gating init}
    \centering
    \begin{adjustbox}{max width=0.8\linewidth}
    \begin{tabular}{ `c  ~c ~c ~c ~c ~c ~c }
        \toprule
        \fmtr{2.35}{$m$}    & \multicolumn{6}{c}{MS-COCO test scores} \\
                               \cmidrule(lr){2-7}
        \null               & B-1  & B-4  & M    & R    & C    & S    \\
        
        \midrule
\rbf    ~~5.0               & 71.6 & 31.4 & 24.6 & 52.8 & 94.4 & 17.5 \\
        ~~2.5               & 71.3 & 31.1 & 24.4 & 52.5 & 93.1 & 17.4 \\
        ~~0.0               & 71.3 & 30.6 & 24.4 & 52.6 & 92.4 & 17.3 \\
        $-$2.5              & 70.8 & 30.1 & 24.1 & 52.1 & 91.1 & 17.0 \\
        $-$5.0              & 70.5 & 29.5 & 23.6 & 51.8 & 88.0 & 16.5 \\
        
        \bottomrule
    \end{tabular}
    \end{adjustbox}
\end{table}

\begin{table}[t]
    \caption{Effects of varying sparsity loss weightage $\lambda_{s}$ with $s_{target} = 0.9$ on SA model.} 
    \label{table: Pruning: Ablation: Sparsity weight}
    \centering
    \begin{adjustbox}{max width=0.9\linewidth}
    \begin{tabular}{ `c  ~c ~c ~c ~c ~c ~c ~c }
        \toprule
        \fmtr{2.35}{$\lambda_{s}$}  & \fmtr{2.35}{\makecell{Sparsity\\(\%)}} & \multicolumn{6}{c}{MS-COCO test scores} \\
                                                                                \cmidrule(lr){3-8}
        \null                       & \null    & B-1  & B-4  & M    & R    & C    & S    \\
        
        \midrule
         1.0                & 66.2     & 71.4 & 31.0 & 24.7 & 52.7 & 94.1 & 17.4 \\
         2.0                & 83.0     & 71.6 & 31.0 & 24.6 & 52.7 & 93.9 & 17.5 \\
   \rbf  5.0                & 90.0     & 71.4 & 30.8 & 24.4 & 52.4 & 93.1 & 17.3 \\
        10.0                & 90.0     & 71.1 & 30.6 & 24.4 & 52.5 & 92.7 & 17.3 \\
        
        \bottomrule
    \end{tabular}
    \end{adjustbox}
\end{table}

Table~\ref{table: Pruning: Ablation: Gating init} shows the effect of different gating initialisation values $m$. From the results, we can establish that the best overall performance is achieved when $m = 5.0$. This can be attributed to the fact that initialisation value of $5.0$ allows model parameters $\theta$ to be retained with high probability at early stages of training, leading to better convergence. This observation is also consistent with the works of \cite{zhu2017prune,narang2017exploring}, where it is found that gradual pruning can lead to better model performance. Thus, we recommend setting $m = 5.0$.

Table~\ref{table: Pruning: Ablation: Sparsity weight} shows the effect of sparsity regularisation weightage $\lambda_{s}$. This is an important hyperparameter that could affect the final sparsity level at convergence, with higher sparsity target $s_{target}$ requiring larger $\lambda_{s}$. From the results, we can see that low values lead to insufficient sparsity. At the same time, we found that setting $\lambda_{s}$ to a large value does not necessarily degrade its final performance, as $\lambda_{s}$ of 80 and 120 were used for UD and ORT models in Fig.~\ref{fig: Pruning: UpDown ORT COCO}. 


\section{Discussion}
\label{sec: Pruning: Discussion}


In the formulation of SMP, the sparsity loss $L_{s}$ is annealed using an inverted cosine curve $\alpha$ defined in Eq.~\eqref{eq: Pruning: sparsity anneal}. This annealing schedule is inspired by works on gradual pruning as well as works on Variational Recurrent Auto-Encoder (VRAE) for text generation. In particular, \cite{narang2017exploring} has found that gradual pruning is 7\% to 9\% better than hard pruning. 
Our experiments that compared gradual-uniform pruning and hard-uniform have found this to be generally true, especially at high sparsity levels (see Sec.~\ref{subsec: Pruning: Experiments: Pruning} and \ref{subsec: Pruning: Experiments: Pruning Sequence}).
Meanwhile, loss annealing is also used to train VRAE for text generation in the form of Kullback-Leibler (KL) annealing \cite{bowman2016generating}. Specifically, the KL regularisation term is gradually introduced during training in order to shift the model from a vanilla RAE to a VRAE.
In the same spirit, SMP gradually transitions the model from dense to sparse, as shown in Fig.~\ref{fig: Pruning: Training progression}.

Another perspective on the effectiveness of gradual pruning or sparsity annealing can be found in the notion of ``Information Plasticity'' introduced by \cite{achille2018critical}. In the work, it is found that DNN optimisation exhibits two distinct learning phases: a critical ``memorisation phase'' during which information stored in the weights as measured by Fisher Information rapidly increase, followed by a ``forgetting phase'' where the amount of information contained gradually decrease and the network is less adaptable to change. This suggests that an ideal pruning schedule should impose sparsity constraints while the network has passed its critical learning phase, and at the same time still plastic enough to adapt to such changes.

Beyond this, sensitivity-based pruning is another equally important aspect of SMP. In SNIP \cite{lee2018snip}, weights are pruned based on the absolute magnitude of the derivative of training loss with respect to the multiplicative pruning masks. In contrast, SMP achieves this by updating the gating parameters according to their gradients. This crucial difference meant that whereas SNIP removes weights with the least influence on training loss regardless of sign, SMP will also remove weights with negative influence (\ie{} increase training loss). 

Moreover, SNIP computes sensitivity at initialisation using one or more batches of training data. This implies that SNIP can be sensitive to the choice of weight initialisation scheme, as stated in the paper. In contrast, SMP performs continuous and gradual sparsification throughout the training process, making it less sensitive to weight initialisation. In fact, Section~\ref{subsec: Pruning: Experiments: Pruning} shows that SMP can be used on a variety of architectures, each with its own set of initialisation schemes.

By combining these insights, we are able to realise several benefits.
Firstly, SMP achieves \emph{good performance across sparsity levels from 80\% to 99.1\%} (111$\times$ reduction in NNZ parameters). This is in contrast with competing methods \cite{zhu2017prune,see2016compression} where there is a significant performance drop-off starting from sparsity level of 90\% (see Sec.~\ref{subsec: Pruning: Experiments: Pruning}).
Secondly, our SMP sparsity loss \emph{allows explicit control of the overall pruning ratio and compression desired} by simply specifying the target sparsity $s_{target}$. The pruning ratio for each layer is also automatically determined (see Fig.~\ref{fig: Pruning: Layerwise sparsity}). In contrast, works like \cite{srinivas2017training,louizos2018learning} control sparsity levels indirectly via a set of regularisation hyperparameters.
Last but not least, SMP can be \emph{easily implemented on top of any model}, and be \emph{integrated seamlessly into a typical training process}. Only 2 main hyperparameters needed to be tuned (gating learning rate and $\lambda_{s}$), instead of up to 4 as in \cite{zhu2017prune,narang2017exploring}. Since pruning is performed in-parallel with training, we can avoid the complexities and costs associated with iterative train-and-prune \cite{dai2020grow} or reinforcement learning techniques \cite{he2018amc}. Complexities associated with variational pruning \cite{chirkova2018bayesian} such as the local reparameterisation trick can also be avoided.


\section{Conclusion and Future Work}
\label{sec: Pruning: Conclusion}

This paper presented empirical results on the effectiveness of unstructured weight pruning methods on various image captioning architectures, including RNN and Transformer architectures.
In addition, we presented an effective end-to-end weight pruning method -- Supermask Pruning -- that performs continuous and gradual sparsification based on parameter sensitivity.
Subsequently, the pruning schemes are extended by adding encoder pruning, where we showed that conducting decoder pruning and training simultaneously provides good performance.
We also demonstrated that using appropriate pruning methods, ideal sparsity levels can be found in the range of 80\% to 95\%. These sparse networks can match or outperform their dense counterparts.
Finally, we show that for a given performance level, a large-sparse LSTM captioning model outperforms a small-dense one in terms of model costs.
In short, this is the first extensive attempt at exploring unstructured model pruning for image captioning.
We hope that this work can spur new research interest in this direction and subsequently serve as benchmark for future image captioning pruning works.

We believe that this work opens up a sea of directions for future works.
Firstly, optimised sparse matrix multiplication kernels and block-sparsity patterns can be implemented in order to realise speed-up at inference time.
Finally, there are many other pruning methods that are yet to be tested, including variational pruning and saliency-based methods.



\bibliographystyle{elsarticle-num}
\bibliography{ref}


\end{document}